\documentclass[preprint,12pt]{elsarticle}

\usepackage{amssymb}
\usepackage{amsthm}
\usepackage{lineno}
\usepackage{bm}
\usepackage[colorlinks=true, pdfauthor=author]{hyperref}
\usepackage{wrapfig}
\usepackage{booktabs} 
\usepackage{algorithm}
\usepackage{algorithmic}
\usepackage{graphicx}
\usepackage{amsmath}
\usepackage{amssymb}
\usepackage{amsfonts}
\usepackage{mathptmx}
\usepackage{multirow}
\usepackage{graphicx}
\usepackage{mathrsfs}
\usepackage{colortbl}
\usepackage{subfigure}
\usepackage{indentfirst}
\usepackage{ulem}
\usepackage{url}
\usepackage{verbatim}
\usepackage{rotating}
\usepackage{adjustbox}
\usepackage{wrapfig}

\graphicspath{{./images/}}

\journal{arXiv}

\begin{document}

\begin{frontmatter}



\title{NCART: Neural Classification and Regression Tree for Tabular Data}


\author[first]{Jiaqi Luo}
\ead{jiaqi.luo@dukekunshan.edu.cn}
\author[first]{Shixin Xu \corref{cor1}}
\cortext[cor1]{Corresponding author}
\ead{shixin.xu@dukekunshan.edu.cn}
\affiliation[first]{organization={Data Science Research Center, Duke Kunshan University},
            addressline={No.8 Duke Ave.}, 
            city={Kunshan},
            postcode={215000}, 
            state={Jiangsu Province},
            country={China}}

\begin{abstract}

Deep learning models have become popular in the analysis of tabular data, as they address the limitations of decision trees and enable valuable applications like semi-supervised learning, online learning, and transfer learning. However, these deep-learning approaches often encounter a trade-off. On one hand, they can be computationally demanding when dealing with large-scale or high-dimensional datasets. On the other hand, they may lack interpretability and may not be suitable for small-scale datasets. In this study, we propose a novel interpretable neural network called Neural Classification and Regression Tree (NCART) to overcome these challenges. NCART is a modified version of Residual Networks that replaces fully-connected layers with multiple differentiable oblivious decision trees. By integrating decision trees into the architecture, NCART maintains its interpretability while benefiting from the end-to-end capabilities of neural networks. The simplicity of the NCART architecture makes it well-suited for datasets of varying sizes and reduces computational costs compared to state-of-the-art deep learning models. Extensive numerical experiments demonstrate the superior performance of NCART compared to existing deep learning models, establishing it as a strong competitor to tree-based models.

\end{abstract}




\end{frontmatter}


\section{Introduction}
\label{s:intro}
Tabular data is ubiquitous across various fields and industries, representing a wide range of valuable information, including financial records, customer information, and scientific measurements. Its pervasive presence underscores the importance of accurate prediction in tabular data analysis. Such predictions empower businesses, researchers, and analysts to fully leverage the potential of these datasets and drive meaningful advancements in their respective domains.

Decision tree models are widely recognized as an effective machine learning technique, particularly well-suited for analyzing tabular data. They offer several advantages, including interpretability, handling both numerical and categorical data, ease of implementation, and scalability.
Decision trees are the basis for more advanced ensemble methods such as random forests \cite{breiman2001random} and gradient boost decision trees (GBDT) \cite{chen2016xgboost,ke2017lightgbm,prokhorenkova2018catboost}.
Nevertheless, they suffer from a lack of flexibility, discontinuity and the inability to pre-trained models, online learning, and transfer learning, which limits their use.

The capabilities of deep learning \cite{lecun2015deep} in extracting high-level features, its end-to-end training capability, and overall flexibility have led to remarkable achievements in various domains, including image analysis \cite{he2016deep}, natural language processing \cite{vaswani2017attention}, and generative modeling \cite{goodfellow2020generative}.
In the context of modeling tabular data, deep learning has emerged as a promising approach to address the limitations of decision trees. Researchers have been actively exploring innovative techniques \cite{borisov2022deep} with the aim of overcoming the drawbacks of traditional methods and expanding the range of applications suitable for deep learning.

Recent studies have highlighted that deep learning models may not always outperform other machine learning methods, such as gradient boosting when applied to tabular data \cite{borisov2022deep, gorishniy2021revisiting, grinsztajn2022tree}. The relatively inferior performance of deep learning models in this context can be attributed to various factors.
One reason is the limited availability of labeled training data. Deep learning models typically require large amounts of labeled data to effectively learn meaningful representations and generalize well. When training data is limited, the performance of deep learning models may be compromised.
Additionally, many deep learning models often contain a large number of parameters. When handling high-dimensional or large-scale datasets, they can be time-consuming and require more resources.
Interpretability remains a significant challenge in deep learning due to the complexity of these models. With their numerous layers and millions of parameters, deep learning models often act as black boxes, making it difficult to understand the decision-making process behind their predictions. This lack of interpretability can be a significant drawback, especially in clinical or financial applications where limited data sets and interpretability are crucial for informed decision-making and interventions.

To tackle these challenges, we propose a novel tree-induced deep learning model called Neural Classification and Regression Tree (NCART). Drawing inspiration from the success of differentiable decision trees \cite{yang2018deep, popov2019neural} and ResNet \cite{gorishniy2021revisiting} in handling tabular data, our model combines the strengths of decision trees and deep learning.
Instead of using fully-connected layers, our model replaces them with a sum of several differentiable decision trees. This allows our model to capture interpretability similar to traditional tree models while benefiting from the end-to-end training capabilities of deep learning. Furthermore, the simplicity of the ResNet architecture ensures that our model achieves desirable results in a limited time.
Our proposed model is designed to work effectively with datasets of varying sizes and demonstrates competitive performance compared to other state-of-the-art machine learning methods. By integrating the superiority of decision trees with the computational power of deep learning, our model not only achieves good performance but also provides insights into the underlying patterns and relationships within tabular data.
Through this innovative combination, our NCRAT model bridges the gap between interpretability and performance in the analysis of tabular data, making it a promising approach for a wide range of applications.

The main contributions of this paper are summarized as follows:
\begin{itemize}
\item Proposal of NCART: We introduce NCART, a deep learning model that integrates differentiable decision trees into a fully-connected ResNet.
The model is applicable to datasets of varying scales while maintaining lower computational costs compared to existing deep learning-based methods.

\item Interpretable Model: NCART exhibits interpretability, generating feature importance similar to those produced by traditional decision tree models. This information is valuable for identifying the most influential features in the decision-making process and understanding the underlying patterns in the data.

\item Superior Performance: Through extensive experimentation, NCART demonstrates superior performance compared to state-of-the-art deep learning models. Furthermore, it remains competitive with decision tree models, showcasing its efficacy in tackling complex tasks.
\end{itemize}

\section{Related Work}
\label{s:rel}
In this section, we provide an overview of the concepts from previous research that are pertinent to deep learning for tabular data. We divide the deep learning modes into three categories according to the network structure, namely tree-induced networks, transformer-based networks, and other specialized models.

\subsection{Tree-induced Neural Networks}
Given the effectiveness of tree-based models for tabular data, researchers have made efforts to incorporate differentiable trees that combine the advantages of tree ensembles with gradient-based optimization of neural networks. Several notable approaches in this direction are summarized as follows:
\begin{itemize}
    \item Deep Neural Decision Trees (DNDT): Proposed by Yang et al. \cite{yang2018deep}, DNDT is a neural network-based tree model that is straightforward to implement. It incorporates a soft binning function for split decisions, enhancing interpretability. However, its scalability is limited in high-dimensional feature spaces due to the Kronecker product computation for binned features.
    \item Neural Oblivious Decision Ensembles (NODE): Inspired by CatBoost, NODE \cite{popov2019neural} employs oblivious decision trees to construct an ensemble model. It is a fully differentiable model that can handle both numerical and categorical features without requiring preprocessing. NODE has demonstrated superior performance compared to GBDT models across multiple datasets, albeit with higher computational costs.
    \item Net-DNF: Net-DNF \cite{katzir2021net} leverages the representation of decision trees as Boolean formulas, specifically disjunctive normal forms (DNF). By utilizing this insight, Net-DNF emulates the characteristics of GBDT and achieves comparable results to XGBoost on certain classification tasks.
    \item Binary Tree Learning: The authors propose a novel method \cite{zantedeschi2021learning} that combines Argmin differentiation and a mixed-integer programming model to simultaneously learn discrete and continuous parameters of the tree, which allows the users to leverage the generalization capabilities of stochastic gradient optimization for decision splits and principled tree pruning. 
\end{itemize}

These approaches aim to combine the interpretability of decision trees with the optimization capabilities of deep learning models, providing promising alternatives for tackling tabular data challenges.

\subsection{Neural Networks based on Transformer}
Inspired by the recent successes of transformer-based models in
natural language processing (NLP) \cite{vaswani2017attention}, researchers have proposed multiple approaches using attention mechanisms for heterogeneous tabular data.
TabNet \cite{arik2021tabnet} is one of the pioneering transformer-based models designed for tabular data. Its architecture consists of multiple subnetworks processed hierarchically in a sequential manner. Each subnetwork represents a decision step that performs soft instance-wise feature selection. The authors demonstrate that TabNet surpasses other variants on various large-scale tabular datasets.
TabTransformer \cite{huang2020tabtransformer} employs self-attention-based transformers to map categorical features to contextual embeddings. This approach enhances robustness to missing or noisy data and enables interpretability. Extensive experiments show that TabTransformer achieves performance comparable to tree-based ensemble techniques, demonstrating success even in the presence of missing or noisy data.
FT-Transformer \cite{gorishniy2021revisiting} is a straightforward adaptation of the Transformer architecture for tabular data. It considers the relationship between numerical and categorical features by feeding them together to transformer blocks, which TabTransformer has not considered.
Self-Attention and Intersample Attention Transformer (SAINT) \cite{somepalli2021saint} is a hybrid attention approach that combines self-attention with inter-sample attention over multiple rows. Additionally, the researchers leverage self-supervised contrastive pre-training to enhance performance for semi-supervised problems. In the original experiments, SAINT consistently outperforms other methods on supervised and semi-supervised tasks.
Non-Parametric Transformer (NPT) \cite{kossen2021self} utilizes the entire dataset as input. By employing attention between data points, the model can model and leverage relations between arbitrary samples for classifying test samples.
However, these transformer-based methods have been observed to incur significant computational costs, including running time and hardware.
TabPFN \cite{hollmann2022tabpfn} is a new transformer-based model that can handle small-scale tabular data in less than a second with no hyperparameters tuning and can get competitive results with GBDT. However, its effectiveness is restricted to classification tasks involving numerical features. Furthermore, performing inference on larger-scale datasets poses a challenge on current consumer GPUs. Recently, inspired by the investigation of inherent characteristics of tabular data pointed out by \cite{grinsztajn2022tree}, researchers have identified that proper handling of feature interactions and feature embedding can enhance the success of neural networks on tabular data. 
ExcelFormer \cite{chen2023excelformer} alternates between two attention modules for feature interactions and embedding updates, employing a specialized training method that gradually optimizes these modules through specific regularization techniques.  What sets ExcelFormer apart is its remarkable performance, as it can achieve results on par with finely tuned models like XGboost and Catboost, all without the need for hyperparameter tuning.
T2G-Former \cite{yan2023t2g} introduces a novel Graph Estimator to organize independent tabular features into a feature relation graph, enabling the development of a specialized tabular learning Transformer designed to compete effectively with GBDT.

\subsection{Models with other specialized structures}
To explain the underlying relationships between features, Zheng et al. introduced Dual-Route Structure-Adaptive Graph Networks (DRSA-Net) \cite{zheng2023deep}. DRSA-Net dynamically learns a sparse graph structure between variables and characterizes interactions through dual-route message passing. Extensive experiments in various application scenarios demonstrate that DRSA-Net is competitive with classical algorithms and deep models.
Regularization Learning Networks (RLN) \cite{shavitt2018regularization} incorporate trainable regularization coefficients for individual weights in a neural network to reduce sensitivity. These coefficients promote sparsity in the network and can outperform deep neural networks, achieving results comparable to those of GBDT. However, the evaluation is based on a dataset primarily composed of numerical data and cannot handle categorical features.
Wide\&Deep \cite{cheng2016wide} and DeepFM \cite{guo2017deepfm} are designed for online prediction where the input space is tabular data. They both use embedding techniques to handle the categorical features, but the difference is that Wide\&Deep employs a linear model and a wide selection of hand-crafted logical expressions on features to improve the generalization capabilities while DeepFM applies a learned Factorization Machines to replace the hand-crafted features to improve the model performance.
The DeepGBM model \cite{ke2019deepgbm} is also for online prediction, it consists of two components: CatNN and GBDT2NN. The former is to handle the categorical features and has the same structure as DeepFM, the latter is to distill knowledge from a trained GBDT. By integrating the advantages of GBDT on tabular data and the flexibility of deep neural networks, DeepGBM can outperform other models on various online prediction problems.
The DCN-V2 \cite{wang2021dcn} framework is introduced as an improved approach to address the limitations of these existing feature interaction learning methods, enhancing practicality in large-scale industrial applications while outperforming state-of-the-art algorithms on benchmark datasets.

The lack of a consistent structure in general tabular data, unlike image and language data, poses a limitation on the application of deep learning. In order to address this challenge, researchers have devised several strategies.
One prominent approach involves the transformation of tabular data into images or text. By utilizing more efficient models in CV and NLP, these efforts aim to enhance the performance of deep learning in the domain of tabular data \cite{sun2019supertml, yin2020tabert}.
Another solution is designing effective neural network architectures that aim to generate high-level features to boost performance, including TabNN \cite{ke2018tabnn}, DANets \cite{chen2022danets}, and TabCaps \cite{chen2022tabcaps}. This innovative approach empowers deep learning to work better with tabular data, making it more flexible and efficient in handling the unique patterns that tabular data poses.
Furthermore, Value Imputation and Mask Estimation (VIME) \cite{yoon2020vime} extends the success of self- and semi-supervised learning in CV and NLP to tabular data.
AutoGluon Tabular \cite{erickson2020autogluon} builds ensembles of basic neural networks together with other traditional ML techniques, with its key contribution being a strong stacking approach.

Most deep neural networks are trained using back-propagation (BP), which may yield suboptimal performance or limited generalization \cite{suganthan2018non}. To overcome this drawback, researchers have utilized Random Vector Functional Link (RVFL) networks \cite{pao1994learning} to design models for tabular data \cite{shi2021random,shi2022weighting}. Experimental results demonstrate that the proposed networks are competitive with other state-of-the-art neural networks. However, the method lacks comparisons with GBDT, which is usually considered the preferred choice for tabular data.

\section{Neural Classification and Regression Tree}
\label{s:method}
In this section,  detailed descriptions of Neural Classification and Regression Tree (NCART) are presented. 
We first introduce the differentiable decision trees. Next, we present the architecture of NCART. Finally, we conclude this section with a discussion on the interpretability of NCART.

\subsection{Approximation of a Decision Tree}
The standard decision tree is a recursive subdivision of the feature space, but this process cannot be made differentiable, and of course, it cannot be expressed by a neural network.
Instead of using the traditional decision tree, we have chosen to utilize the Oblivious Decision Tree (ODT) \cite{rokach2005decision}. The idea is inspired by the successful endeavor to make the ODT differentiable, as previously mentioned \cite{popov2019neural}.

An oblivious decision tree is a type of binary decision tree where each level tests the same feature.
It splits the data along each feature and compares each feature to a learned threshold.
The leaf value of an ODT is defined as
\begin{equation}
\label{eq.odt}
    c(\mathbf{x}) = g(H(x_1-s_1), \cdots, H(x_n-s_n)),
\end{equation}
where $\mathbf{x}=(x_1,\cdots,x_n) \in \mathbb{R}^n$ is the input feature vector, $\mathbf{s}=(s_1,\cdots,s_n) \in \mathbb{R}^n$ is the learned threshold,  $g: \mathbb{R}^n \to \mathbb{R}$ is a function mapping each split region to its corresponding leaf value and
\begin{equation}
    H(x) = \begin{cases} 0, & \text{if } x < 0 \\ 1, & \text{if } x \geq 0 \end{cases} 
\end{equation}
is the \textit{Heaviside step function} that indicates the decision routes.

However, an ODT does not consider the interaction between different features when making splits, which makes it less expressive compared to standard trees.
To address the issue, we combine multiple ODTs together to approximate the representation ability of a standard decision tree.

\begin{figure*}[htp]
\centering
\subfigure{\includegraphics[scale=0.4]{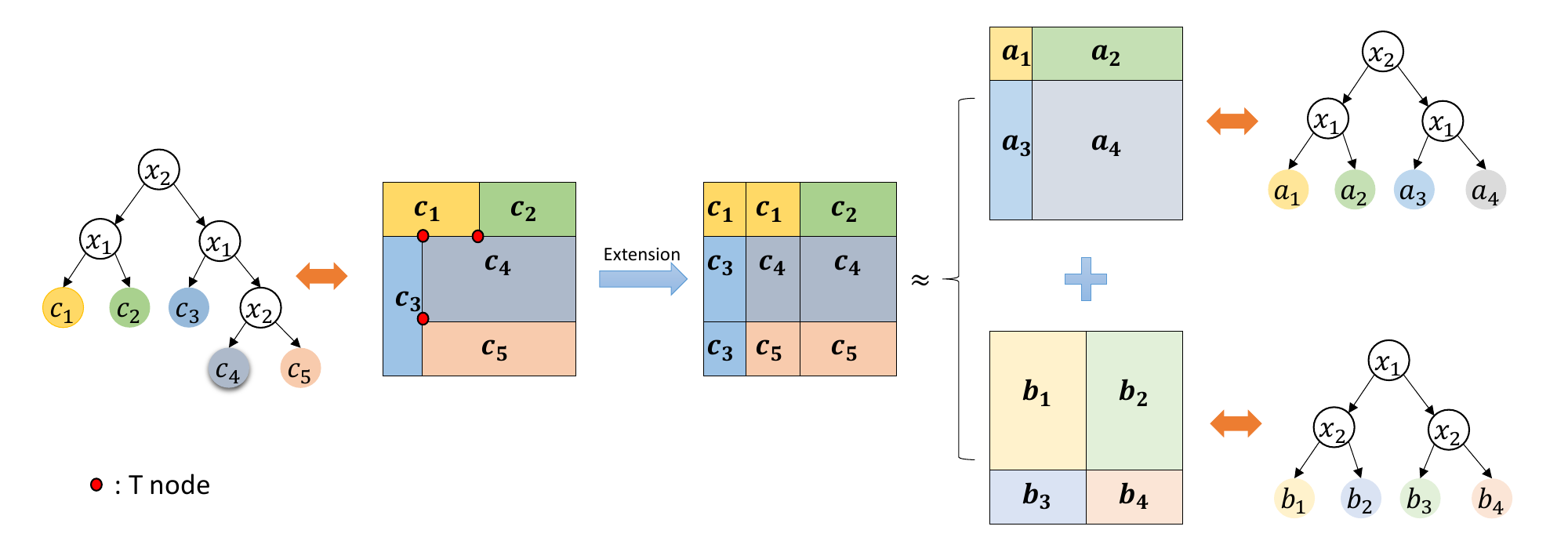}}
\caption{Illustration of approximation of} a decision tree using two oblivious trees. The leaf values have the following relationships: $a_1+b_1\approx c_1$, $a_2+b_1\approx c_1$, $a_2+b_2\approx c_2$, $a_3+b_1\approx c_3$, $a_3+b_3\approx c_3$, $a_4+b_1\approx c_4$, $a_4+b_2\approx c_4$, $a_4+b_3\approx c_5$, $a_4+b_4\approx c_5$.
\label{f.approx}
\end{figure*}

Figure \ref{f.approx} presents an example to illustrate how to approximate a decision tree with two ODTs. 
On the left, there is a decision tree that partitions the feature space into a mesh comprising 5 distinct subregions. Within this mesh, there are nodes characterized by three connected edges, resulting in a grid-like structure resembling the letter "T".
To further refine the representation, we extend the connections for each of these "T" nodes, creating a grid mesh. It's important to note that this extension does not impact the value of each individual region.
Subsequently, we can approximate the refined feature space by considering it as the sum of two separate spaces. Each space has 4 distinct subregions and it can be effectively expressed as ODTs.
The leaf values of the ODTs can be determined by solving the following objective:
\begin{align*}
   & \min |ODT_1+ODT_2-DT| \\
    = &\min_{a_i, b_j}\left\{|a_1+b_1-c_1|+|a_2+b_1-c_1|+|a_2+b_2-c_2|\right.\\
    &+|a_3+b_1-c_3| + |a_3+b_3-c_3|+|a_4+b_1-c_4|\\
    &\left.+|a_4+b_2-c_4|+|a_4+b_3-c_5| +|a_4+b_4-c_5|\right\}, 
\end{align*}
 where $ ODT_i$ and $DT$ are the vectors formed by the leaf values of ODT  and the decision tree, respectively. 
By solving this optimization problem, we can effectively approximate a decision tree using two distinct oblivious trees.

It is important to highlight that ensemble tree models can be approximated using ODTs as they essentially consist of models composed of multiple trees. However, to achieve similar performance, ODTs often require a larger number of trees compared to standard decision trees.
Furthermore, in an ODT, while nodes at the same level share the same test feature, the split values may vary. Since the results are based on an ensemble of ODTs, we can focus on a specific formulation where each level of ODTs has identical split values. This approach helps reduce the dimensionality of the function and simplifies the computation process.

\subsection{NCART Architecture}
\begin{figure*}[htp]
\centering
\subfigure{\includegraphics[scale=0.5]{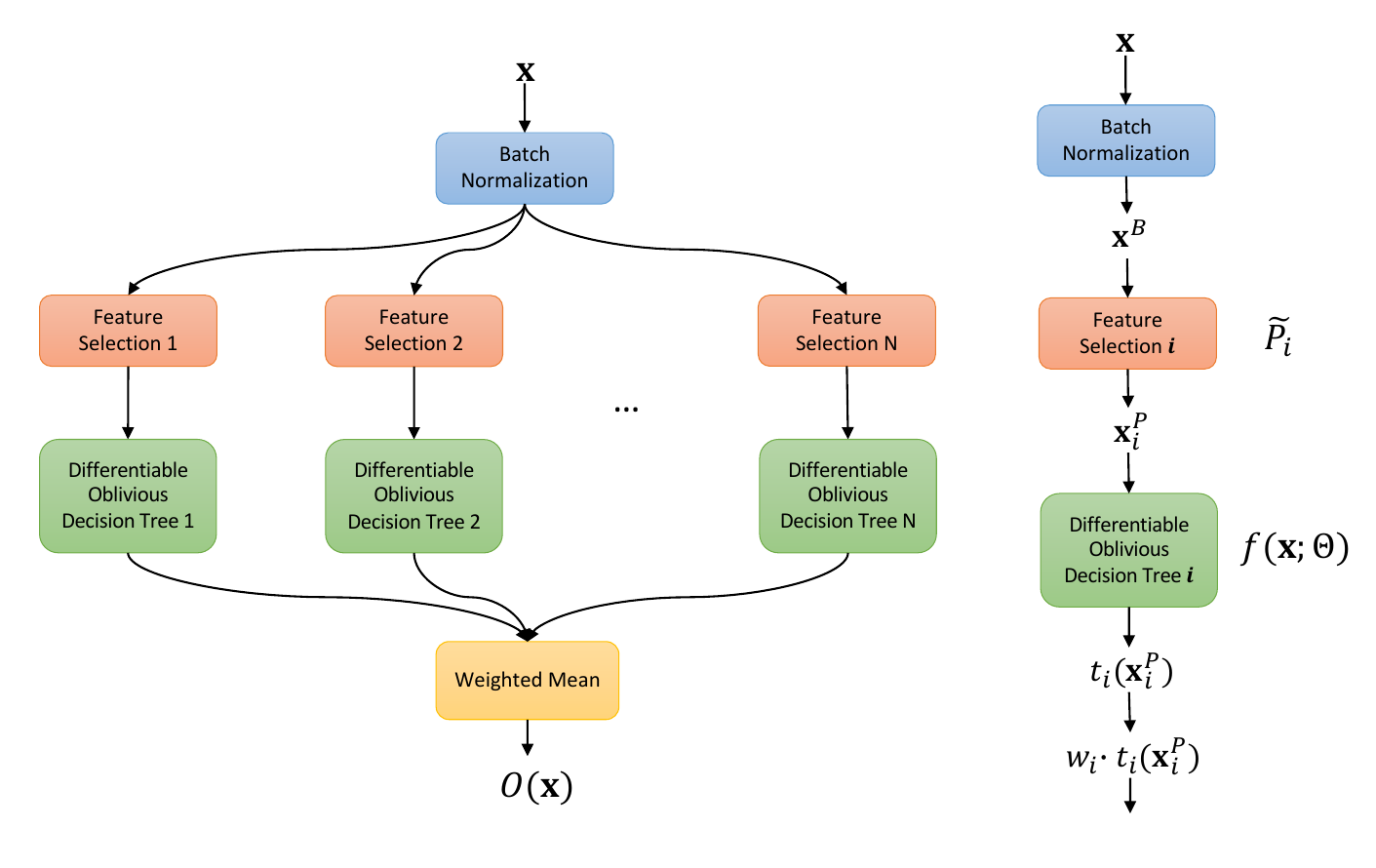}}
\caption{Structure of an NCART block with 4 components: Batch Normalization (blue block) for data preprocessing, Feature Selection (red block, Eq. \eqref{e.feat_sel}), Differentiable Oblivious Tree (green block, Eq. \eqref{e.ndt}), and Weight Mean for ensemble (yellow block, Eq. \eqref{e.ncart}).}
\label{f.block}
\end{figure*}

The architecture of an NCART block, which is illustrated in Figure \ref{f.block}, consists of 4 components: Data preprocessing, Feature Selection, Differentiable Oblivious Trees, and Ensemble.

\paragraph{Data preprocessing}
Suppose the input is $\mathbf{x}$, we do not do any special processing on the input vector (including categorical features), and the vector is directly sent to the first layer which is a simple \textit{Batch Normalization}:
\begin{equation}
\label{e.bn}
\mathbf{x}^B = Batch Normalization(\mathbf{x}).
\end{equation}

\paragraph{Feature selection}
Feature selection plays a critical role in enhancing model performance by identifying and incorporating the most influential features. In an ideal scenario, our objective is to ascertain a projection matrix, denoted as $\mathbf{P}_{d,n}$, with the stipulation that each of its rows corresponds to a one-hot vector:
\begin{equation}
\label{e.feat_sel}
\widetilde{\mathbf{x}} = \mathbf{P}\mathbf{x}^B.
\end{equation}
Here, $\mathbf{x}^B = (x_{1}^B,\cdots, x_{n}^B)^\intercal$, $\widetilde{\mathbf{x}} = (x_{k_1}^B,\cdots, x_{k_d}^B)^\intercal$, and ${k_1, k_2, \cdots, k_d}$ represents a subsequence of ${1, \cdots, n}$. The elements of matrix $\mathbf{P}$, denoted as $P_{ij}$, are binary, taking values in the set ${0,1}$, and they adhere to the constraint $\sum_{j=1}^{n} P_{ij} = 1$ for all $i$.

However, it is hard to learn $\mathbf{P}$ due to its discrete nature, existing in a discrete integer space. In our research, we introduce an innovative approach that involves the application of a sparse transformation to each row of a learnable matrix. This process leads to the creation of a sparse projection, which is mathematically represented as:
\begin{equation}
\label{e.sel_mat}
\widetilde{\mathbf{P}} = h(\mathbf{A}) = \begin{bmatrix}
h(\mathbf{A}_1) \\
h(\mathbf{A}_2) \\
\cdots\\
h(\mathbf{A}_d)
\end{bmatrix}.
\end{equation}
Specifically, $\mathbf{A}$ is a learnable matrix in the neural network. $\mathbf{A}_i=(A_{i1}, \cdots, A_{in})$ is the $i_{th}$ row of $\mathbf{A}$. $h$ is a sparse function similar to the softmax, but able to output sparse probabilities, ensuring that the sum of each row remains equal to 1.  One popular choice for the sparse function is the \textit{Sparsemax} \cite{martins2016softmax} or \textit{entmax} \cite{peters-etal-2019-sparse}, which is considered a hyperparameter in our network configuration. Further details regarding the mathematical expressions defining the sparse function can be found in \ref{a.equation}.

Subsequently, we apply this sparse projection to the feature vector $\mathbf{x}$, resulting in the projected feature vector $\mathbf{x}_P$:
\begin{equation}
\label{e.feat_sel2}
\mathbf{x}^P = \widetilde{\mathbf{P}}\mathbf{x}^B.
\end{equation}

\paragraph{Differentiable Oblivious Decision Tree}
In order to make \eqref{eq.odt} differentiable, we undertake the following two modifications:
\begin{itemize}
    \item We replace the non-differentiable function $H(x)$ with the sigmoid function, denoted as $\sigma(x)=\frac{1}{1+e^{-x}}$. 
    \item We change $g$  to be a two-layer fully-connected network $f$ with $ReLU$ activation, which is expressed as:
    \begin{equation}
    \label{e.dnn}
    f(\mathbf{x}; \Theta) = \mathbf{W_2}(ReLU(\mathbf{W_1}\mathbf{x}+\mathbf{b_1})) + \mathbf{b_2},
    \end{equation}
    where $\Theta$ are the parameters in the network.
\end{itemize}

Subsequently, these modifications yield a differentiable version of the ODT,
which is defined as:
\begin{equation}
\label{e.ndt}
\begin{aligned}
t(\mathbf{x}^P) &= f(\sigma(\mathbf{x}^P-\mathbf{s}; \Theta)) \\
&= f(\sigma(x_1^P-s_1), \sigma(x_2^P-s_2), \dots, \sigma(x_n^P-s_n); \Theta).
\end{aligned}
\end{equation}
The key advantage of these modifications is that the differentiability of both the sigmoid function $\sigma$ and the function $f$ enables the tree model $t$ to also be differentiable, thereby enhancing its capacity for learning.

\paragraph{Ensemble}

All the individual outputs are collectively directed to the final layer, where a weighted vote is conducted to produce the output, defined as:
\begin{equation}
\label{e.ncart}
O(\mathbf{x}) = \frac{1}{N}\sum_{i=1}^{N}w_i\cdot t_i(\mathbf{x}_i^P)
\end{equation}
Here, $\mathbf{x}_i^P = \widetilde{\mathbf{P}}_i\mathbf{x}^B$ is the output of $i_{th}$ projection (see Fig. \ref{f.block}). The parameters $w_i$, with $i$ ranging from 1 to $N$, are considered to be learnable, allowing the model to adapt and optimize these parameters during the training process.

\paragraph{NCART Network Architecture}

In a GBDT algorithm, each new tree is employed to learn the residual errors or gradients of the loss function. A similar concept is adopted in the ResNet, where each new block represents the residual mapping to be learned, effectively behaving like an ensemble of shallow networks~\cite{veit2016residual}.

\begin{figure}[htp]
\centering
\includegraphics[scale=0.45]{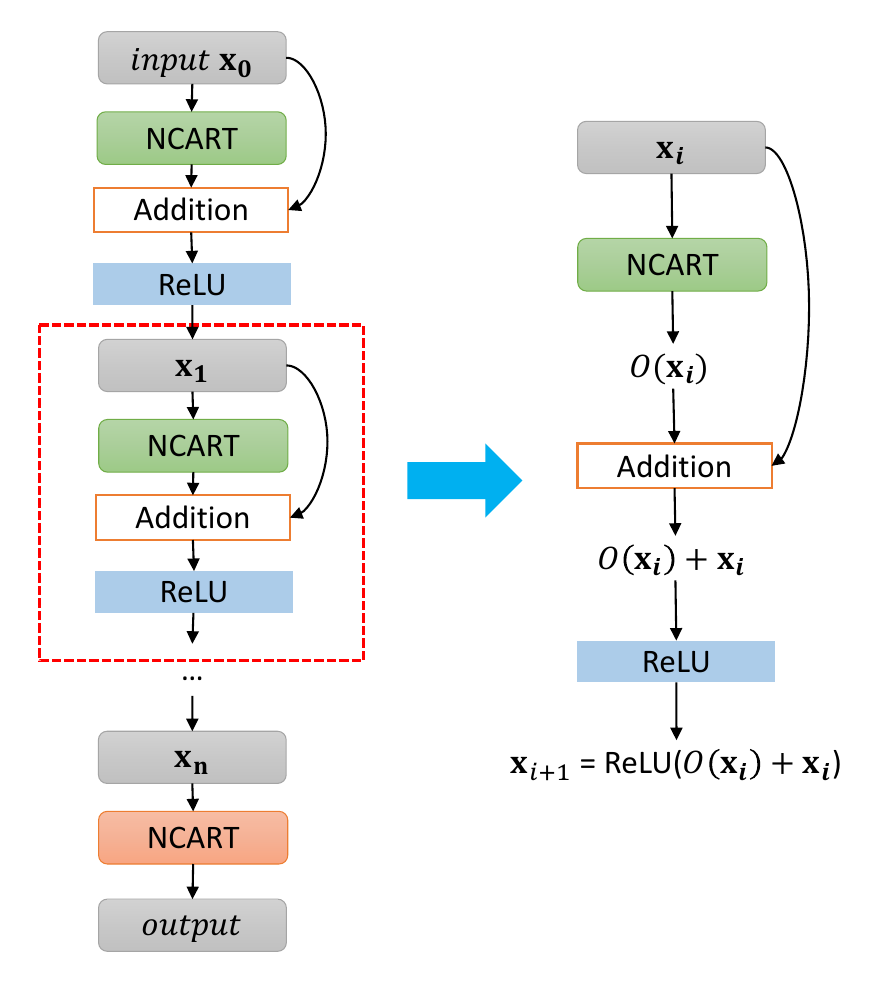}
\caption{NCART Network architecture. A green-filled block indicates the utilization of all features for feature splitting, while a red-filled block signifies the presence of a feature selection layer.}
\label{f.resblock}
\end{figure}

Drawing inspiration from these techniques, we propose to combine multiple NCART blocks to enhance the overall model's performance. The architecture of our ensemble model is illustrated in Fig.~\ref{f.resblock}, resembling a modified feed-forward ResNet in which each fully-connected block is replaced with an NCART block.
In our default configuration, each NCART block consists of an equal number of differentiable trees, and the final NCART block, highlighted in red, serves the purpose of dimensionality reduction.

\subsection{Feature Importance}
Calculating feature importance in NCART involves a process similar to that of traditional decision trees. Since our model consists of multiple tree structures, the methodology for determining feature importance remains akin to conventional decision tree analysis.

For a differentiable ODT, we can easily derive the decision rule for a feature from Eq.\eqref{e.ndt}. For a given feature index $j$, if $\sigma(x_j-s_j) > 0.5$, it indicates that $x_j > s_j$ and the corresponding samples will be assigned to the right leaf. Conversely, if $\sigma(x_j-s_j) \le 0.5$, the samples will be assigned to the left leaf.
With this approach, we can directly apply the Gini coefficient, commonly used in traditional decision trees, to measure the impurity of feature $x_j$. The formulation for the Gini coefficient is as follows: 
\begin{equation}
\label{e.feat_import1}
   I_j = 1-\left(\frac{m_{j}^{left}}{M}\right)^2-\left(\frac{m_{j}^{right}}{M}\right)^2,
\end{equation}
where $M$ is the number of samples. $m_{j}^{left}$ is the number of samples divided into the left leaf, and $m_{j}^{right}$ is the number of samples that are divided into the right leaf.

If the NCART block includes a feature selection layer, the Gini coefficient can be formulated as follows:
\begin{equation}
\label{e.feat_import2}
  I_j^{sel} = \sum_{k=1}^{d} \left(1-\left(\frac{m_{k}^{left}}{M}\right)^2-\left(\frac{m_{k}^{right}}{M}\right)^2\right) h(A)_{kj}.
\end{equation}
where $d$ is the dimensionality of feature selection.

In an NCART with $L$ blocks (Fig.~\ref{f.resblock}), where the last block contains a feature selection layer, each differentiable tree is independent, and every feature in a differentiable tree only splits once. Thus, to calculate the Gini coefficient of a feature $x_j$ in an NCART, we can obtain it by summing up the results from all the individual trees:
\begin{equation}
\label{e.feat_import3}
   Importance_j = \sum_{l=1}^{L-1}\sum_{k=1}^{N} I_{lkj} + \sum_{k=1}^{N} I_{kj}^{sel},
\end{equation}
where $I_{lkj}$ represents the Gini coefficient of feature $x_j$ at  the $k_{th}$ tree of the $l_{th}$ layer.



\section{Experiments}
\label{s:exps}
\subsection{Experiments Setup}
\paragraph{Datasets} 
We conduct comparisons on 20 datasets, which are all from OpenML\footnote{\url{https://www.openml.org/}}. 
These datasets serve diverse purposes, including both classification and regression tasks. They encompass a range of sample sizes and feature types, combining categorical and numerical variables.
For all datasets, we didn't do any special preprocessing, except apply the ordinal encoding to the categorical features and specify the corresponding indices for CatBoost and Transformer-based models. More details about the datasets can be found in \ref{a.data}.

\paragraph{Benchmark models}

Our comparative analysis includes the proposed models and benchmarking against three GBDT models and seven widely-used deep learning baselines: XGBoost \cite{chen2016xgboost}, CatBoost \cite{prokhorenkova2018catboost}, LightGBM \cite{ke2017lightgbm}, NODE \cite{popov2019neural}, TabNet \cite{arik2021tabnet}, SAINT \cite{somepalli2021saint}, FT-Transformer \cite{gorishniy2021revisiting}, TabCaps \cite{chen2022tabcaps}, RLN \cite{shavitt2018regularization}, and ResNet \cite{gorishniy2021revisiting}. It's important to note that TabCaps is exclusively designed for classification tasks and may not be suitable for regression tasks.

\paragraph{Evaluation metrics} 
For classification tasks, both binary and multi-class, we assess model performance using the \textbf{AUC} (Area Under the Receiver Operating Characteristic Curve). Additionally, we employ the \textbf{F1-score} as an evaluation metric for classification tasks, considering potential dataset imbalances. In the case of regression tasks, we utilize the \textbf{MSE} (Mean Squared Error) to evaluate the performance of the various methods.

\paragraph{Training procedure} 
Our training procedure follows the guidelines outlined in \cite{borisov2022deep} for all baseline models. We evaluate the performance of our models using their open-source code\footnote{\url{https://github.com/kathrinse/TabSurvey}}. To tune hyperparameters, we utilize the Optuna library\footnote{\url{https://optuna.org/}} with 10 trials for each model. Each trial configuration undergoes 5-fold cross-validation. Detailed descriptions of the adjusted hyperparameters can be found in \ref{a:params}, while other parameters remain at their default values.
For neural network models, we conduct training for 1000 epochs using the Adam optimizer\cite{kingma2014adam} with a learning rate of 0.001. Our training batches consist of 1024 samples, with validation batches comprising 256 samples.
For each algorithm and each 5-fold cross-validation, we run the algorithm for up to 50000s to prevent excessively long runtimes.
All experiments are conducted on a workstation equipped with an Intel Core i9-10900X CPU, 128GB RAM, and an NVIDIA-3080 GPU.

\subsection{Performances}
\begin{sidewaystable}
\renewcommand\arraystretch{1.3}
    \centering
    \begin{adjustbox}{width=\textwidth}
    \begin{tabular}{lcccccccccccc}
    \toprule[2pt]
        \multirow{2}{*}{Dataset} & \multirow{2}{*}{Metric} & \multicolumn{11}{c}{Model} \\
         &  & XGBoost & CatBoost & LightGBM & NODE & TabNet & SAINT & FTTrans & TabCaps &RLN & ResNet & NCART \\
    \midrule[1.5pt]
    \multicolumn{13}{c}{Binary Classification}\\

        \multirow{2}{*}{diabetes}& AUC &82.44 $\pm$ 4.03 &83.53 $\pm$ 3.24 & 82.63 $\pm$ 3.5 &82.98 $\pm$ 3.11 &54.06 $\pm$ 1.79& 62.30$\pm$ 5.63& 81.64 $\pm$ 3.39& 74.55 $\pm$ 5.39 & 63.59$\pm$ 6.21 &82.66 $\pm$ 3.5 & \underline{\textbf{83.72$\pm$ 3.39}}\\
        
        & F1-score & 59.69$\pm$4.25 &   60.9 $\pm$6.46&    44.19$\pm$23.19 &  62.51$\pm$6.62 &  39.17$\pm$2.01 &  12.17$\pm$17.34 &         57.71$\pm$4.83 &   50.04$\pm$9.33 &  47.65$\pm$5.56 &  62.06$\pm$4.47 & \underline{\textbf{ 64.61$\pm$5.39}}\\
        \midrule[0.2pt]

        \multirow{2}{*}{credit-g} &AUC &78.14 $\pm$ 4.19 &78.45 $\pm$ 6.03 &\underline{\textbf{79.13$\pm$4.31}}  &76.66 $\pm$4.63 &57.34 $\pm$ 1.98 &77.54 $\pm$ 5.0& 77.38$\pm$6.2& 74.79$\pm$0.94 &  54.22 $\pm$ 6.31&74.67$\pm$4.03 & 76.86$\pm$2.74\\
        
        & F1-score &   83.62$\pm$2.4 &  \underline{\textbf{84.53$\pm$2.03}} &    83.22$\pm$2.56 &  82.69$\pm$1.67 &  81.89$\pm$0.28 &  83.43$\pm$2.47 &  82.95$\pm$2.98 &  83.09$\pm$1.00 &  82.53$\pm$0.35 &  80.35$\pm$2.77 &   84.2$\pm$1.79\\
        \midrule[0.2pt]

        \multirow{2}{*}{qsar-biodeg} &AUC &93.43 $\pm$1.52 &92.87$\pm$2.02 & 93.54 $\pm$1.15&92.48$\pm$2.04 &57.34 $\pm$5.89 &93.01$\pm$1.09 & 92.88$\pm$2.14& 93.01$\pm$2.19   &91.33$\pm$1.71 & 92.93$\pm$1.38 & \underline{\textbf{93.64$\pm$1.14}}\\
        
        & F1-score &    80.1$\pm$3.74 &    79.42$\pm$3.00 &    80.42$\pm$1.52 &  79.89$\pm$1.82 &  42.18$\pm$4.75 &  80.56$\pm$3.18 &         78.43$\pm$3.18 &   81.96$\pm$2.81 &   75.3$\pm$1.38 &  81.13$\pm$2.8 & \underline{\textbf{ 82.14$\pm$1.71}}\\
        \midrule[0.2pt]

        \multirow{2}{*}{scene} &AUC &99.14$\pm$0.19  &99.21$\pm$0.39 & 99.30$\pm$0.26 &98.93$\pm$0.20  &81.26$\pm$2.79 & $OOM$ & $OOM$ & 98.96$\pm$0.54& \underline{\textbf{99.30$\pm$0.16}}&  98.06$\pm$1.63 & 98.65$\pm$0.36\\
        
        & F1-score &   94.76$\pm$1.69 &    96.88$\pm$1.20 &    94.36$\pm$1.74 &   95.1$\pm$1.64 &  70.01$\pm$4.58 &  $OOM$ & $OOM$ &    \underline{\textbf{97.00$\pm$1.34}} &  96.88$\pm$1.20 &  94.99$\pm$1.33 &  95.27$\pm$0.62\\
        \midrule[0.2pt]

        \multirow{2}{*}{ozone-level} &AUC &91.58$\pm$2.51  &91.87$\pm$2.94 & 91.94$\pm$2.80 &89.97$\pm$3.40  &67.13$\pm$5.93 & 89.18$\pm$1.28& 85.94$\pm$2.59& 89.80$\pm$9.28 &  64.14$\pm$13.39& 89.35$\pm$4.02& \underline{\textbf{93.10$\pm$2.04}}\\
        
        & F1-score &   \underline{\textbf{36.99$\pm$10.15}} &    28.71$\pm$11.61 &    36.94$\pm$14.65 &    0.0$\pm$0.00 &  19.11$\pm$10.23 &   26.0$\pm$8.82 &         21.93$\pm$15.38 &     26.38$\pm$17.24 &   4.03$\pm$5.61 &  13.14$\pm$2.02 &  27.15$\pm$15.12\\
        \midrule[0.2pt]

        \multirow{2}{*}{delta$\_$ailerons} &AUC &97.91$\pm$0.36  &97.85$\pm$0.34 & \underline{\textbf{97.98$\pm$0.22}} &97.38$\pm$0.29  &97.03$\pm$1.14 & 96.38$\pm$0.39& 97.48$\pm$0.42& 96.86$\pm$0.64 &  60.67 $\pm$ 18.47 & 97.18$\pm$0.48 & 97.56$\pm$0.38\\
        
        & F1-score &  \underline{\textbf{94.88$\pm$0.20}} &    94.68$\pm$0.29 &    94.78$\pm$0.29 &  94.42$\pm$0.33 &  89.73$\pm$10.22 &  93.88$\pm$0.46 &         94.52$\pm$0.38 &   94.16$\pm$0.54 &  69.34$\pm$0.02 &  92.52$\pm$1.90 &  94.69$\pm$0.46 \\
        \midrule[0.2pt]

        \multirow{2}{*}{MagicTelescop} &AUC &93.46$\pm$0.56  &\underline{\textbf{93.57$\pm$0.75}} & 93.47$\pm$0.67 &92.05$\pm$0.60  &92.71$\pm$0.86 & 90.23$\pm$1.66& 93.16$\pm$0.57& 92.69$\pm$0.60 &  84.96$\pm$1.7& 90.80$\pm$1.99& 92.98$\pm$0.86\\
        
        & F1-score &   85.55$\pm$0.86 &  \underline{\textbf{  85.72$\pm$1.16 }} &    85.54$\pm$1.08 &  84.08$\pm$0.40 &  84.14$\pm$0.57 &  81.52$\pm$1.51 &   84.63$\pm$0.41 &   84.63$\pm$1.14 &  76.65$\pm$1.15 &  81.29$\pm$3.07 &  84.73$\pm$1.51\\
        \midrule[0.2pt]

        \multirow{2}{*}{jannis} &AUC &87.34$\pm$0.14  &\underline{\textbf{87.52$\pm$0.22}} &87.13$\pm$0.23 &84.65$\pm$0.29  &86.17$\pm$0.38 & 78.39$\pm$1.29 & 87.02$\pm$0.24& 85.58$\pm$0.27 &  80.38$\pm$1.36 & 85.58$\pm$0.30 & 86.74$\pm$0.16\\
        
        & F1-score &   80.09$\pm$0.11 &   \underline{\textbf{ 80.38$\pm$0.25}} &    79.93$\pm$0.18 &  77.92$\pm$0.40 &  79.03$\pm$0.57 &  72.66$\pm$1.51 &         80.17$\pm$0.41 &   78.62$\pm$0.30 &  72.04$\pm$1.61 &  78.48$\pm$0.39 &  79.54$\pm$0.33\\
        \midrule[0.2pt]
        
        \multirow{2}{*}{road-safety} &AUC &\underline{\textbf{90.24$\pm$0.29}}  &89.29$\pm$0.27 &90.11$\pm$0.28 &87.43$\pm$0.87  &88.54$\pm$0.24 & 86.62$\pm$0.87& 76.14$\pm$0.24& 84.56$\pm$1.89 &  54.78$\pm$2.99 & 79.53$\pm$0.49 & 88.23$\pm$0.18\\
        
        & F1-score &  \underline{\textbf{ 81.86$\pm$0.33 }} &    80.55$\pm$0.32 &    81.76$\pm$0.34 &  78.75$\pm$0.40 &  79.42$\pm$0.29 &  77.44$\pm$0.42 &         69.78$\pm$0.92 &   75.16$\pm$1.15 &  27.58$\pm$32.89 &  72.42$\pm$0.37 &  79.17$\pm$0.23\\
        \midrule[0.2pt]

        \multirow{2}{*}{Higgs} &AUC  &82.80$\pm$0.05  &82.79$\pm$0.03 & 82.46$\pm$0.10 &82.34$\pm$0.57 & 83.26$\pm$0.43 & 83.62$\pm$0.06& 82.82$\pm$0.17& 82.42$\pm$0.23&   69.9$\pm$2.28 & \underline{\textbf{84.35$\pm$0.05}} & 83.45$\pm$0.09\\
        
        & F1-score &   74.67$\pm$0.08 &    74.68$\pm$0.04 &    74.45$\pm$0.05 &  74.56$\pm$0.22 &  75.04$\pm$0.50 &  75.36$\pm$0.18 &         72.41$\pm$1.02 &   74.61$\pm$0.22 &  63.87$\pm$1.09 &  \underline{\textbf{75.85$\pm$0.19}} &  75.15$\pm$0.20\\

    \midrule[1.5pt]
    \multicolumn{13}{c}{Multi-class Classification}\\

        \multirow{2}{*}{autoUniv-au7} &AUC  &71.43$\pm$ 2.61 & 69.56$\pm$2.80 &\underline{\textbf{72.34$\pm$3.04}} &66.12$\pm$2.79  &51.07$\pm$3.54 & 56.71$\pm$4.55& 62.20$\pm$2.57& 63.72$\pm$3.33 &50.07$\pm$0.14 &58.90$\pm$4.43 & 65.40$\pm$2.30\\
        
        & F1-score &   \underline{\textbf{53.38$\pm$2.61}} &    47.65$\pm$2.80 &    51.68$\pm$3.04 &  44.85$\pm$2.79 &  31.32$\pm$3.54 &  36.22$\pm$4.55 &         38.64$\pm$2.57 &   42.69$\pm$3.33 &  17.52$\pm$0.14 &  39.93$\pm$4.43 &  47.45$\pm$2.30\\
        \midrule[0.2pt]

        \multirow{2}{*}{plants-margin} &AUC  &99.41$\pm$0.10  &\underline{\textbf{99.82$\pm$0.02}} &99.06$\pm$0.18 &99.53$\pm$0.08  &98.37$\pm$0.27 & 99.47$\pm$0.08& 99.46$\pm$0.16& 99.60$\pm$0.10 & 91.64$\pm$0.65 & 99.68$\pm$0.06 & 99.54$\pm$0.05\\
        
        & F1-score &    73.6$\pm$0.10 &  \underline{ \textbf{ 84.43$\pm$0.02}} &    74.53$\pm$0.18 &  78.82$\pm$0.08 &  56.59$\pm$0.27 &  69.96$\pm$0.08 &         72.27$\pm$0.16 &   83.64$\pm$0.10 &   4.45$\pm$0.65 &   80.0$\pm$0.06 &  76.04$\pm$0.05\\
        \midrule[0.2pt]

        \multirow{2}{*}{waveform} &AUC  &96.87$\pm$0.22  &97.09$\pm$0.28 & 96.78$\pm$0.22  &97.26$\pm$0.20  &96.14$\pm$0.45 & 97.09$\pm$0.24& \underline{\textbf{97.42$\pm$0.28}} & 97.25$\pm$0.21 &  97.33$\pm$0.20 & 97.42$\pm$0.34 & 97.17$\pm$0.25\\
        
        & F1-score &   85.42$\pm$0.22 &    86.03$\pm$0.28 &    84.86$\pm$0.22 &  86.48$\pm$0.20 &  84.38$\pm$0.45 &  86.15$\pm$0.24 &         86.89$\pm$0.28 &   86.27$\pm$0.21 &   \underline{\textbf{87.00$\pm$0.20}} &  86.68$\pm$0.24 &  86.34$\pm$0.25\\
        \midrule[0.2pt]

        \multirow{2}{*}{gas-drift} &AUC  &99.95$\pm$0.04 &99.96$\pm$0.03  & 99.96$\pm$0.04 & 99.86$\pm$0.06  &99.88$\pm$0.05 & \underline{\textbf{99.96$\pm$0.02}}& 99.74$\pm$0.09& 99.76$\pm$0.09 & 83.20$\pm$14.58  & 99.88$\pm$0.02 & 99.96$\pm$0.03 \\
        
        & F1-score &   99.44$\pm$0.04 &    99.42$\pm$0.03 &    99.41$\pm$0.04 &  98.43$\pm$0.06 &   98.4$\pm$0.05 &  99.37$\pm$0.02 &         96.16$\pm$0.09 &   98.86$\pm$0.09 &  59.05$\pm$14.58 &  98.83$\pm$0.02 &  \underline{\textbf{99.44$\pm$0.03}}\\
        \midrule[0.2pt]
        
        \multirow{2}{*}{EMNIST} &AUC  &99.49$\pm$0.01  &99.56$\pm$0.01 &99.44$\pm$0.01 &99.42$\pm$0.02  &99.32$\pm$0.03 & $OOM$& $OOM$ & 99.00$\pm$0.03 &98.51$\pm$0.22& \underline{\textbf{99.65$\pm$0.01}} & 99.56$\pm$0.02\\
        
        & F1-score &   83.53$\pm$0.01 &    84.09$\pm$0.01 &    83.18$\pm$0.01 &  80.99$\pm$0.02 &  80.36$\pm$0.03 & $OOM$& $OOM$ &   81.02$\pm$0.03 &   69.9$\pm$0.22 &  \underline{\textbf{85.59$\pm$0.01}} &  84.23$\pm$0.02\\

    \midrule[1.5pt]
    \multirow{2}{*}{Average Rank} &AUC& 3.67 & \underline{\textbf{2.93}} & 3.33 & 5.80 & 7.73 & 7.27 & 6.53 & 6.27 & 8.87 & 5.33 & 3.60 \\
    &F1-score& 3.73 & 3.27 & 4.93 & 6.33 & 8.27 & 7.40 & 7.33 & 5.33 & 9.00 & 5.80 & \underline{\textbf{3.20}} \\
    \midrule[0.2pt]

    \multirow{2}{*}{Best/Worst} &AUC & 1/0 & \underline{\textbf{3/0}}& \underline{\textbf{3/0}} & 0/0 & 0/1 & 1/2 &1/2 & 0/0  &1/8 & 2/0 &\underline{\textbf{3/0}}\\
    &F1-score& \underline{\textbf{4/0}} & 3/0& 0/0 & 0/1 & 0/2 & 0/3 &0/2 & 1/0  &1/6 & 2/1 &3/0\\
    \bottomrule[2pt]
    \end{tabular}
    \end{adjustbox}
    \caption{Mean$\pm$std. results of 11 models on different classification datasets. The \underline{\textbf{bold}} indicates the top result; $OOM$ represents there exists GPU overflow.}
    \label{T.clfresults}
\end{sidewaystable}

\begin{figure}[!ht]
\centering
\subfigure[AUC rank]{\includegraphics[width=0.47\textwidth]{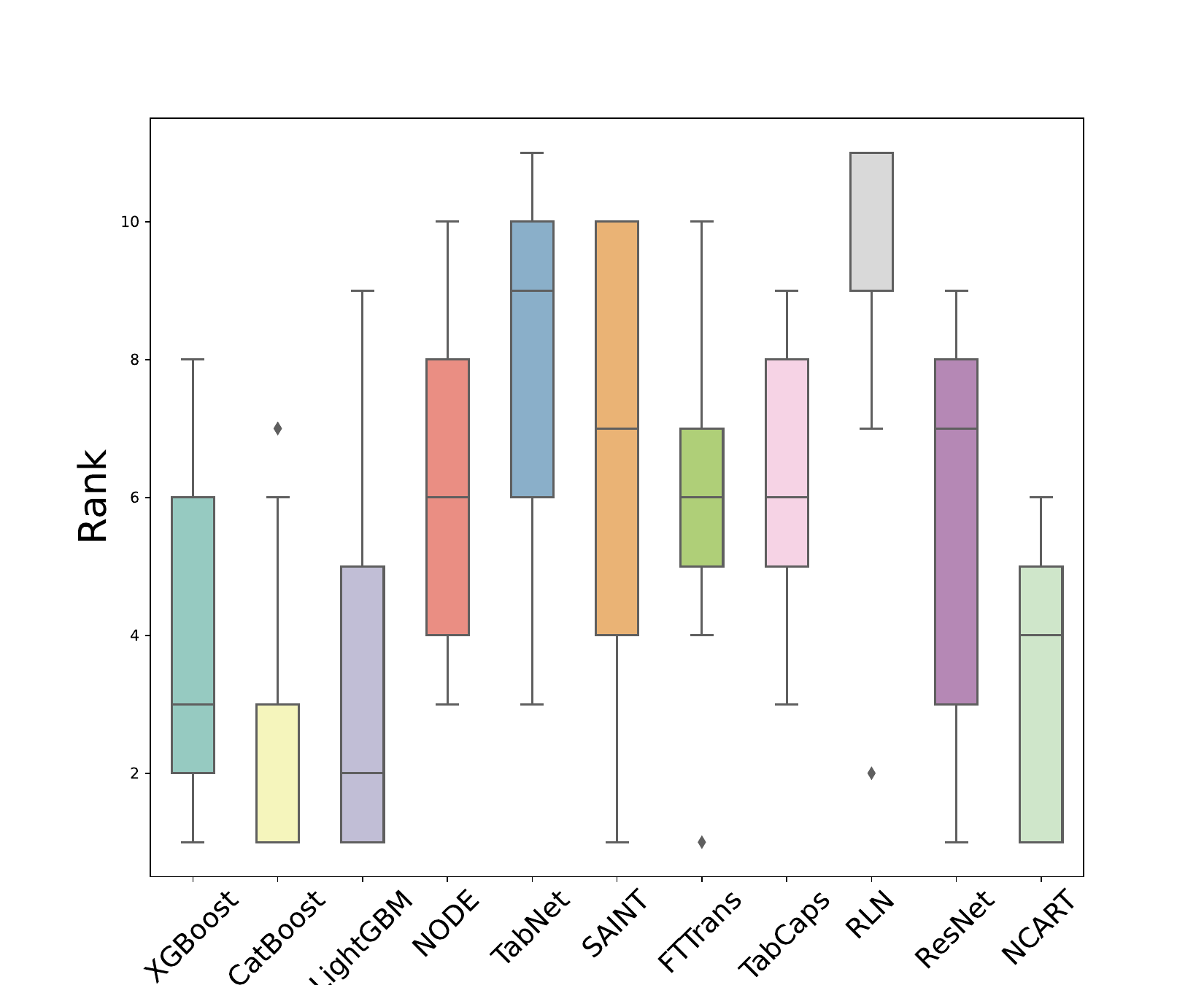}}
\subfigure[F1-score rank]{\includegraphics[width=0.47\textwidth]{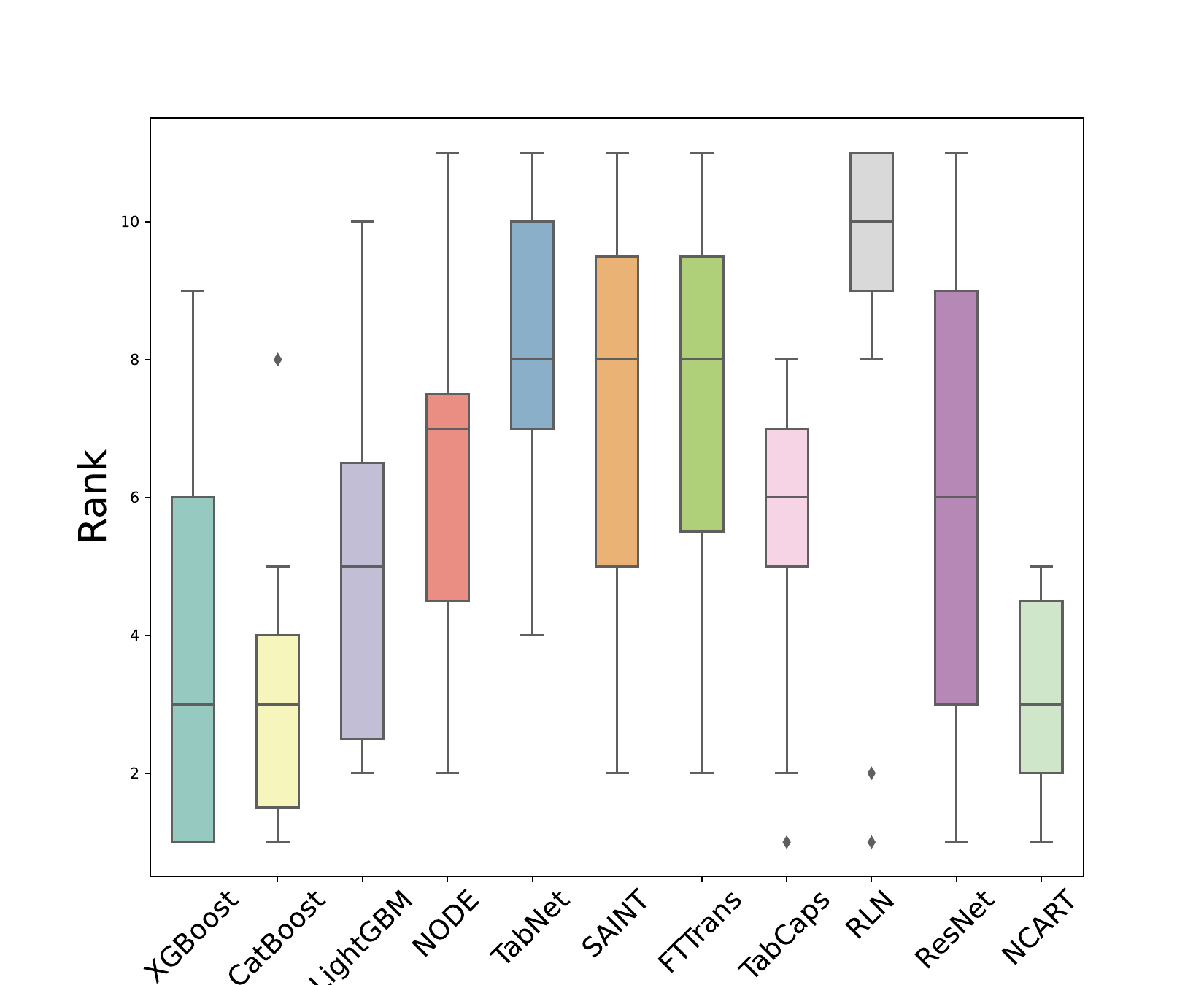}}
\subfigure[MSE rank]{\includegraphics[width=0.47\textwidth]{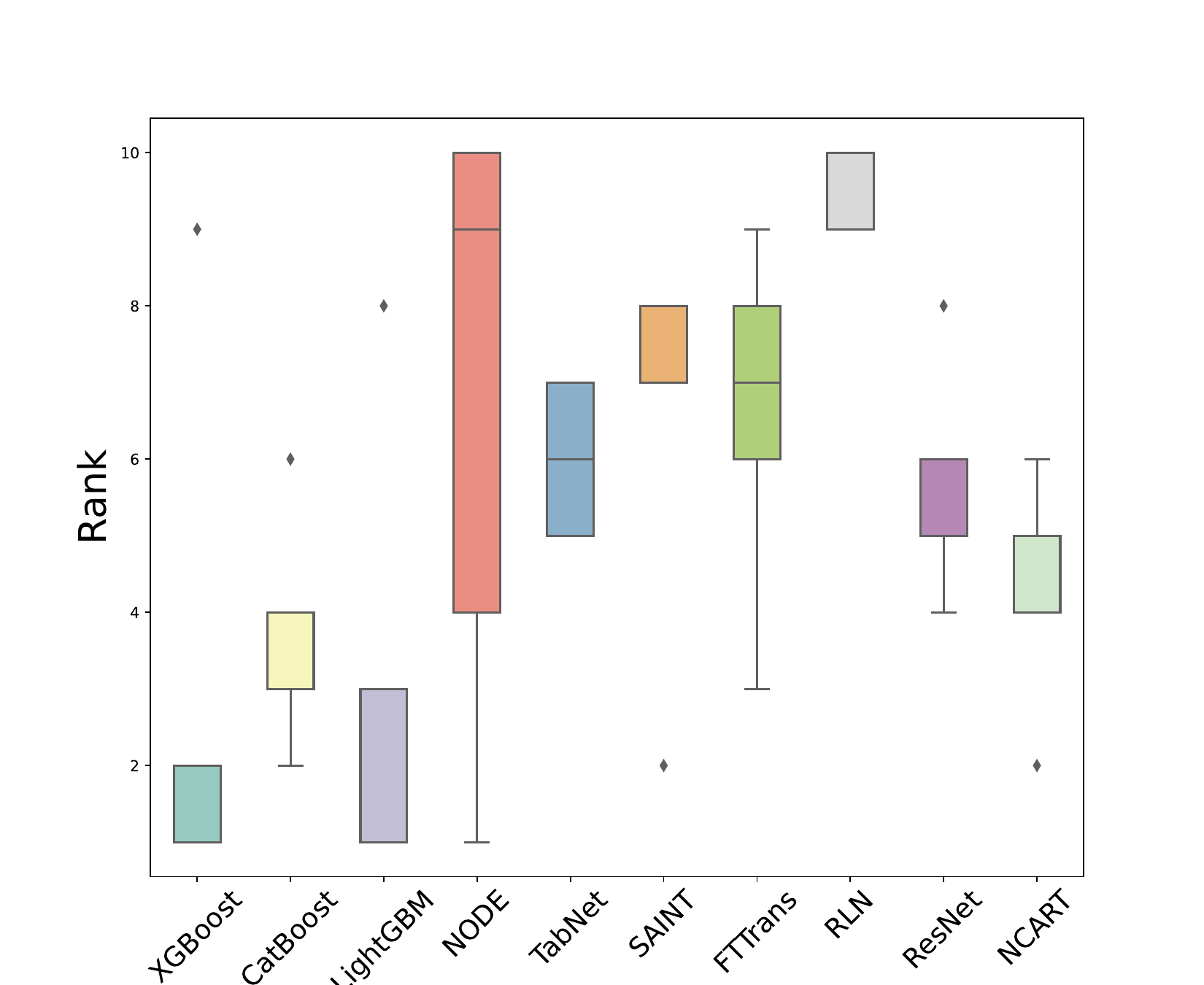}}
\caption{Rank values of different models on 20 datasets. Fig. (a) and (b) are for classification tasks and Fig. (c) is for regression tasks.}
\label{f.rank_results}
\end{figure}

To ensure reproducibility and mitigate model variance, for each approach and dataset, we report the mean and standard deviation out of 5-fold runs for the best configuration.
The results of the comparison are summarized in Table.~\ref{T.clfresults}, Table.~\ref{T.regresults}, and Fig.~\ref{f.rank_results}.

Referring to Table~\ref{T.clfresults}, Table~\ref{T.regresults}, and Figure~\ref{f.rank_results}, it is evident that tree-based models retain their advantages. 
When we look at the AUC, CatBoost stands out with the best average performance. LightGBM and NCART also perform well, achieving the same number of top results, although their average ranks are slightly higher than CatBoost's.
In contrast, a shift in focus towards the MSE highlights XGBoost's superiority, both in terms of best score and average performance.
Furthermore, when considering performance through F1-score, NCART emerges as the top-performing model in terms of average performance. In addition to its elevated average performance, NCART competes strongly with XGBoost in achieving optimal results based on the F1-score.

NODE and TabCaps demonstrate competitive or superior performance on the majority of datasets. NODE excels in terms of the AUC, while TabCaps performs better in F1-score.
Nevertheless, it's important to note that NODE's performance diminishes notably when confronted with high-dimensional regression datasets, like the superconduct dataset. Additionally, when handling the year dataset, which is large-scale and high-dimensional, NODE necessitates significantly longer training time.
TabNet has proven to be a competitive model in cases where the sampling number is large, which matches the highlight in the original paper. However, its performance decreases considerably when the sample size is small. Therefore, its effectiveness can be limited based on the size of the dataset.
SAINT struggles to effectively handle small-scale datasets and requires additional GPU memory when confronted with high-dimensional features, as observed in the context of datasets such as scene and EMNIST. 
Simultaneously, FT-Transformer demonstrates competitive performance, yet it also requires more resources to deal with high-dimensional data.
Although RLN is suitable for all kinds of datasets and can obtain the two best results, it also produces some of the worst scores, indicating a higher variability in their results.
ResNet proves to be a highly effective baseline, particularly when dealing with large-scale datasets. 
NCART, a variant of fully-connected ResNet, surpasses ResNet's performance across a wide range of datasets and delivers superior results comparable to those achieved by other deep learning models.

Additionally, our experiments indicate that deep learning models can achieve comparable or competitive results in classification tasks across datasets of varying sizes. However, when it comes to regression tasks, tree-based methods consistently demonstrate superior performance. 
This discrepancy can be attributed to the discrete nature of tabular datasets and the non-smooth characteristics of the target function. On the contrary, deep learning models tend to be biased towards excessively smooth solutions, which may hinder their performance in regression tasks \cite{grinsztajn2022tree}.

To further substantiate this conclusion, we employ two distinct datasets. The first, analcatdata, comprises categorical features and features only 10 distinct target values, rendering its target distribution discrete. This discrete nature makes it more amenable to tree models, as they excel at approximating piecewise constant functions. In contrast, kin8nm, a dataset stemming from a realistic simulation of forward dynamics, features a continuous target distribution due to the continuous nature of dynamics. Consequently, deep learning models tend to outperform tree-based models on this dataset.

\begin{table*}[!ht]
\renewcommand\arraystretch{1.5}
    \centering
    \begin{adjustbox}{width=\textwidth}
    \begin{tabular}{l|ccccc|cc}
    \toprule[2pt]
        Model &  analcatdata  & kin8nm & superconduct & house$\_$sales & year &Average Rank& Best/Worst\\
    \midrule[1.5pt]
        
        XGBoost  &\underline{\textbf{0.0051$\pm$0.0018}}  &0.0137$\pm$0.0003 & 86.19$\pm$1.71 &0.0281$\pm$0.0012 &\underline{\textbf{74.76$\pm$0.60}} & \underline{\textbf{3.0}} & \underline{\textbf{2/0}} \\

        CatBoost &0.0052$\pm$0.0017  &0.0092$\pm$0.0002 & 105.70$\pm$7.61 &0.0286$\pm$0.0015 &76.74$\pm$0.65 &3.6& 0/0  \\

        LightGBM &0.0054$\pm$0.0018   &0.0111$\pm$0.0004 &\underline{\textbf{85.17$\pm$1.72}}  &\underline{\textbf{0.0281$\pm$0.0008}}& 76.60$\pm$0.55  & 3.2 & \underline{\textbf{2/0}}\\
        
        NODE  &0.0091$\pm$0.0017  &\underline{\textbf{0.0042$\pm$0.0001}} & 1350.31$\pm$39.79 &0.1327$\pm$0.0050&$TimeOut$ & 6.8 & 1/2 \\

        TabNet &0.0273$\pm$0.0161  &0.0100$\pm$0.0014 & 141.46$\pm$10.60 &0.0556$\pm$0.0119&82.03$\pm$4.49 & 6.0 & 0/0\\

        SAINT &0.2988$\pm$0.0129  &0.0044$\pm$0.0001 & 199.16$\pm$4.32 &0.1248$\pm$0.0038&94.12$\pm$0.99  & 6.4 & 0/0\\

        FTTrans &0.3034$\pm$0.0121  &0.0049$\pm$0.0004 &225.70$\pm$12.77 &0.1058$\pm$0.0041&86.37$\pm$2.12 & 6.6 & 0/0 \\

        RLN &0.3219$\pm$0.0434  &0.0347$\pm$0.0050 & 315.75$\pm$44.06 &0.3676$\pm$0.1020& 125.81$\pm$1.04  & 9.6 & 0/3\\

        ResNet &0.0126$\pm$0.0060  &0.0063$\pm$0.0004 & 157.02$\pm$3.93 &0.1295 $\pm$0.0042&77.09$\pm$1.06  & 5.6 & 0/0\\

        NCART &0.0140$\pm$0.0026  &0.0080$\pm$0.0007 & 125.81$\pm$10.66&0.0371$\pm$0.0020&75.06$\pm$0.84 & 4.2 & 0/0\\

    \bottomrule[2pt]
    \end{tabular}
    \end{adjustbox}
    \caption{Mean$\pm$std. results of 10 models on different regression datasets. The \underline{\textbf{bold}} indicates the top result; $TimeOut$ means the running time exceeds the time limit.}
    \label{T.regresults}

\end{table*}

\subsection{Inference time}

\begin{figure}[H]
\centering
\includegraphics[width=0.95\textwidth]{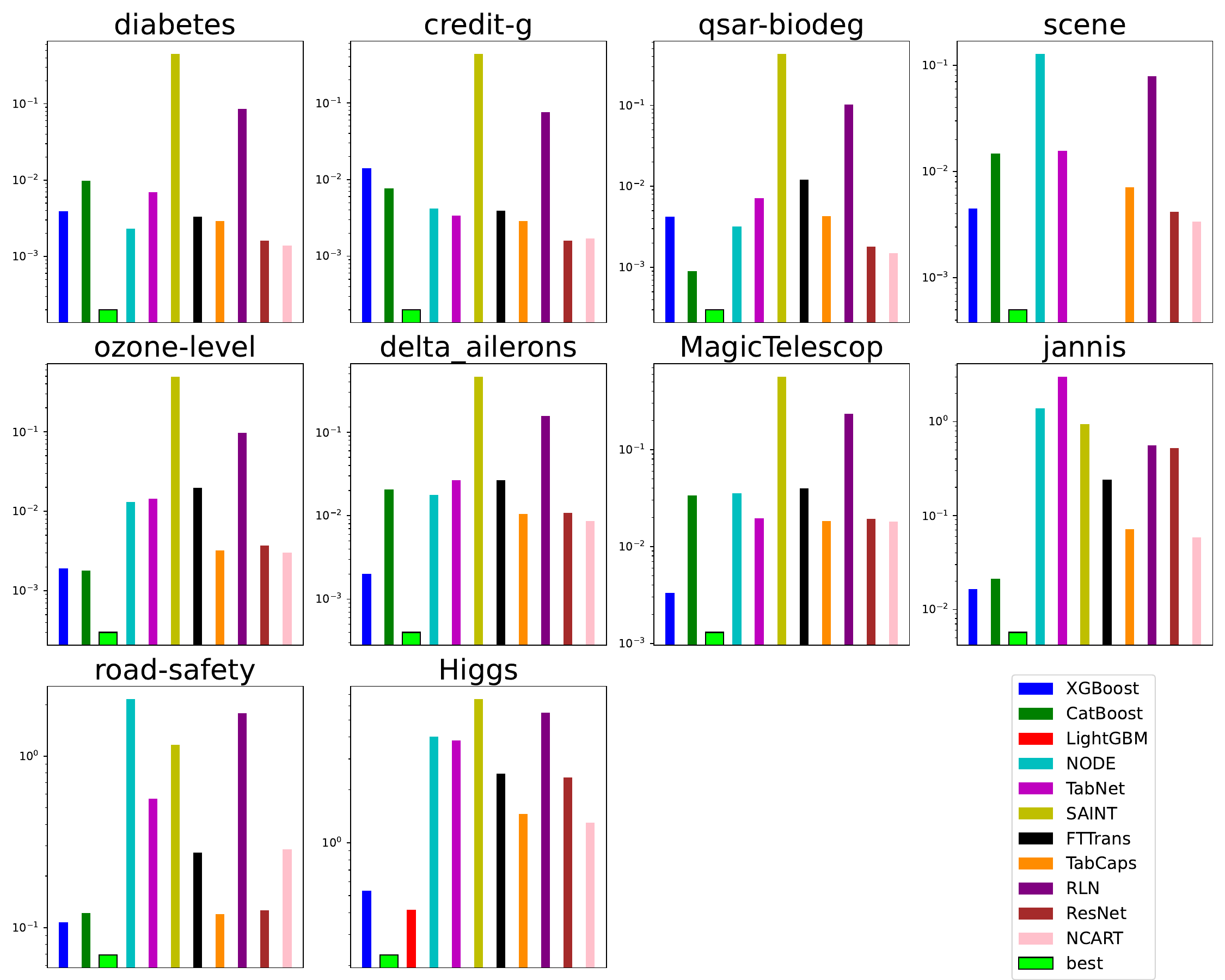}
\caption{Average inference time(s) of a five cross-validation with the best hyperparameters for binary classification tasks. Subfigures are sorted by dataset size. The value on the y-axis represents the inference time. Missing areas indicate that the model has GPU memory overflow.}
\label{f.binary_infer}
\end{figure}

\begin{figure}[!ht]
\centering
\includegraphics[width=0.75\textwidth]{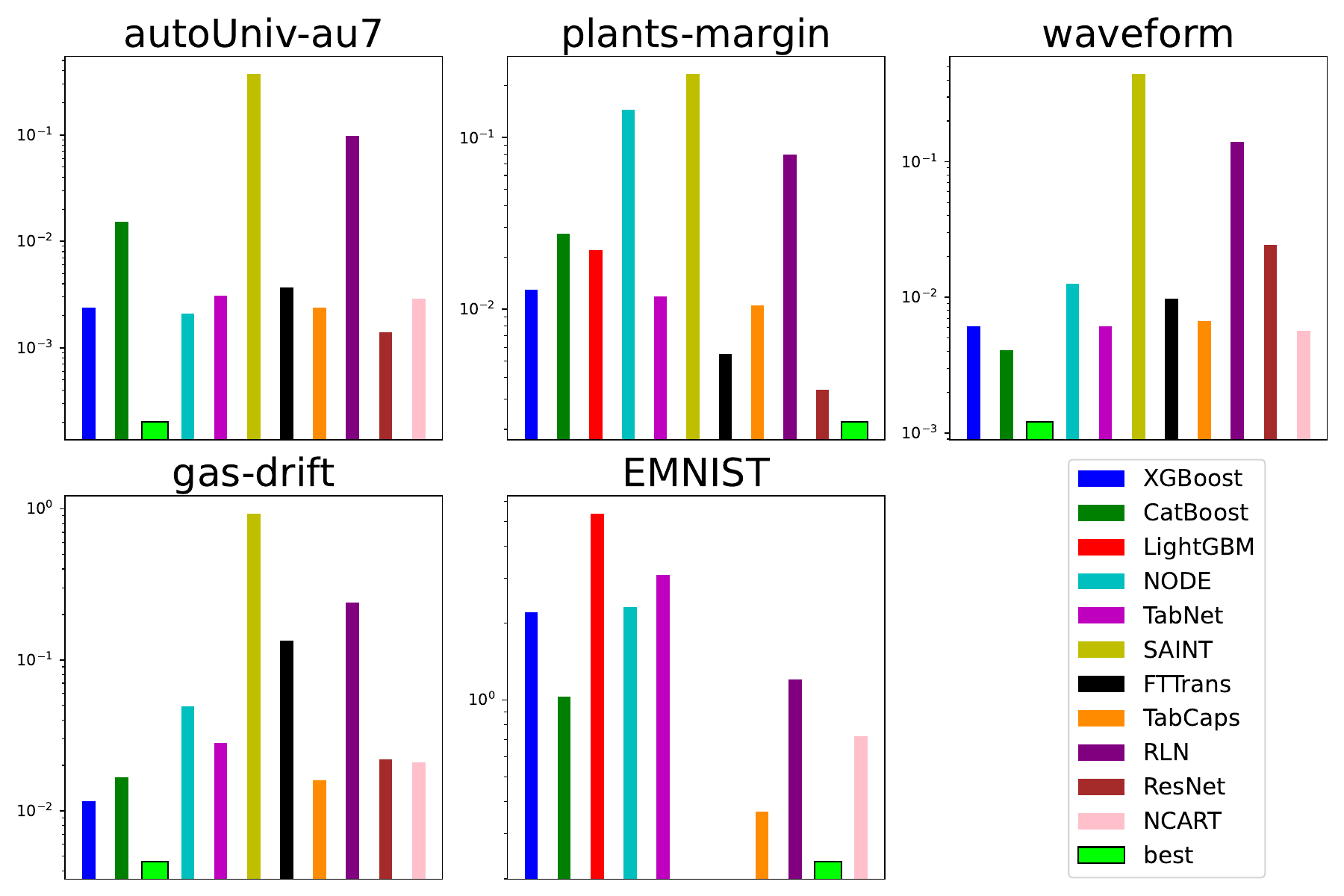}
\caption{Average inference time(s) of a five cross-validation with the best hyperparameters for multi-class classification tasks. Subfigures are sorted by dataset size. The value on the y-axis represents the inference time. Missing areas indicate that the model has GPU memory overflow.}
\label{f.multi_infer}
\end{figure}

\begin{figure}[!ht]
\centering
\includegraphics[width=0.75\textwidth]{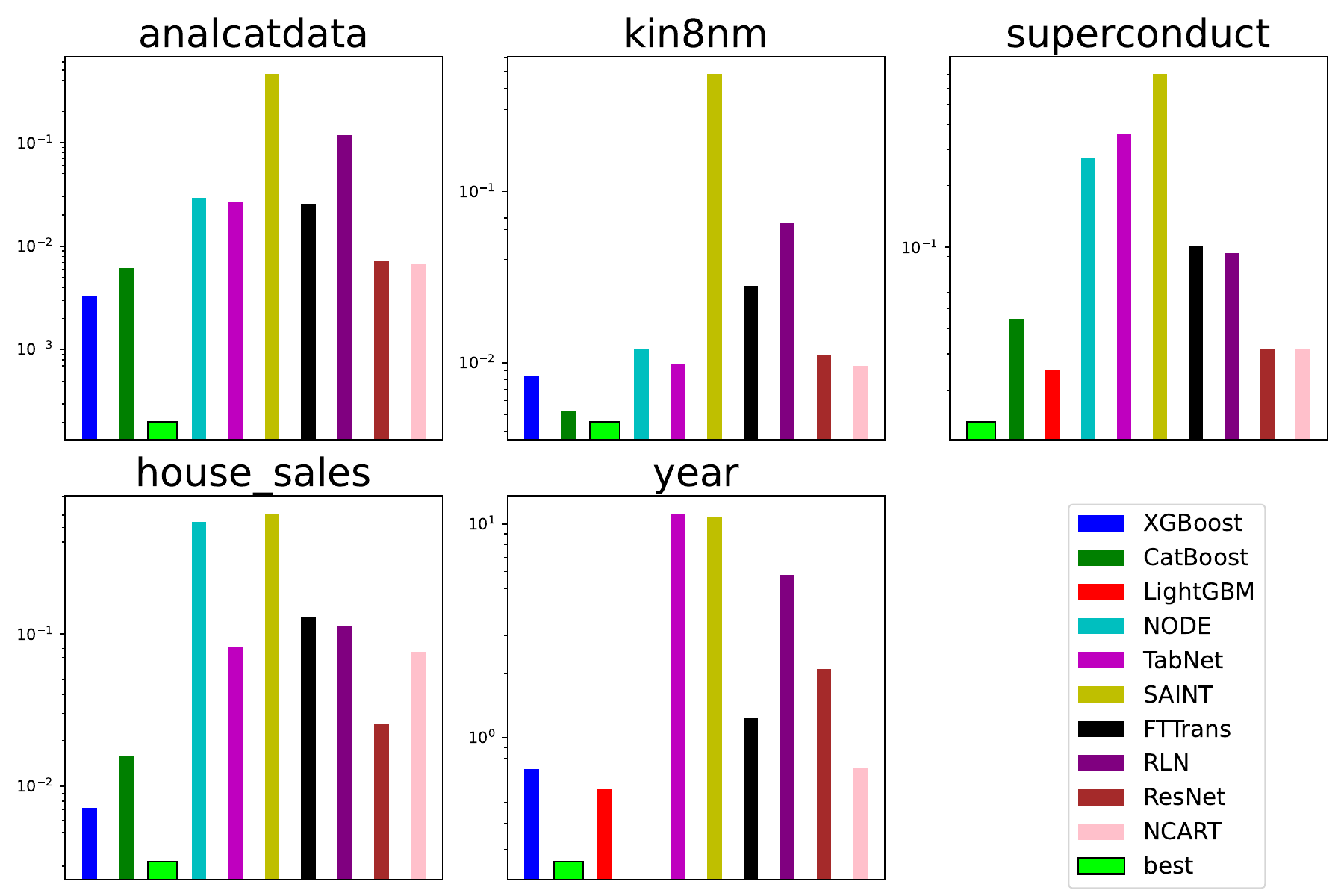}
\caption{Average inference time(s) of a five cross-validation with the best hyperparameters for regression tasks. Subfigures are sorted by dataset size. The value on the y-axis represents the inference time. Missing areas indicate that the model's running time exceeds the time limit.}
\label{f.reg_infer}
\end{figure}

We also analyze the average inference time, which is given in Fig.~\ref{f.binary_infer}, Fig.~\ref{f.multi_infer}, Fig.~\ref{f.reg_infer}, and Table.~\ref{T.infer_time}, of the 5-fold cross-validation with the best parameters of different models in comparison to their performance. Detailed results are presented in \ref{a:infertime}. Besides, we also present the training time in \ref{a:traintime}

Compared to evaluation scores, tree models have bigger advantages in inference time.
When working with a small dataset, most deep learning models produce similar results, except for the SAINT model. And they can outperform XGBoost and CatBoost. However, as the dataset size increases, the computational cost of various models, including NODE, TabNet, SAINT, and RLN, escalates rapidly.
Among deep learning models, TabCaps is the most efficient model, boasting the fastest total computational time for classification tasks.
Although NCART demonstrates slightly lower performance compared to TabCaps, it maintains the smallest average rank among deep learning models for both classification and regression tasks.
ResNet and FT-Transformer also deliver exceptional results in terms of total running time, but their average ranks fall short of competitiveness.

It is indeed surprising that deep learning models exhibit better inference times compared to tree-based methods on two classification datasets, plants-margin and EMNIST, which each contain more than 40 object classes. This is because current GBDT methods are designed for single output, if they want to handle multiple outputs, they need to construct multiple decision trees first and then concatenate the predictions of all trees to obtain multiple outputs. When the output dimension is high, this strategy is not efficient \cite{zhang2020gbdt}. 
Instead, neural networks are more flexible and they just need to adjust the number of neurons in the last layer to adapt to any output dimension. For networks with numerous parameters, such small modification has little impact on the inference time. The computation time of deep learning is more affected by the network structure and dataset size. The structure determines the speed of convergence, while the amount of data influences the number of iterations needed in the training process. These two examples highlight the potential of deep learning models to efficiently handle complex classification tasks with large numbers of classes.

\begin{table*}[!ht]
\renewcommand\arraystretch{1.4}
    \centering
    \begin{adjustbox}{width=\textwidth}
    \begin{tabular}{lccccccccccc}
    \toprule[2pt]
        Dataset & XGBoost & CatBoost &LightGBM & NODE & TabNet & SAINT& FTTrans & TabCaps& RLN & ResNet &  NCART \\
    \midrule[1.5pt]
    \multicolumn{12}{c}{Classification}\\
    Average Rank & 4.27 & 5.07 & \underline{\textbf{2.07}} & 7.47 & 7.27 & 10.4 & 7.6 & 4.4 & 9.67 & 4.4 & 3.47 \\

    Best/Worst& 0/0 & 1/0& \underline{\textbf{12/0}} & 0/0 & 0/1 & 0/8 & 0/2 & 0/0 & 0/5 & 1/0 & 1/0\\

    Total time & 2.9278   &  \underline{\textbf{1.5584}} & 5.8944   & 10.2912    & 10.6376  & $\gg$14.3193  &  $\gg$4.0526   &  2.0984   &  6.0161   & 3.3365 & 2.4406 \\
    \midrule[1pt]

    \multicolumn{12}{c}{Regression}\\

    Average Rank & 2.2 & 2.8 & \underline{\textbf{1.4}} & 8.4 & 6.6 & 9.6 & 7.0 &-&  7.6 & 5.2 & 4.2 \\

    Best/Worst& \underline{\textbf{8/0}} & 1/0& 7/0 & 0/9 & 3/3 & 0/8 & 0/2 &-& 1/0 & 0/0 & 0/0\\

    Total time & 0.7450   &  \underline{\textbf{0.3352}} & 0.6097   &  $\gg$0.8593    & 11.6860  &  13.0290  &  1.5171   &-&  6.1394  &  2.1789   &  0.8495 \\
    \bottomrule[2pt]
    \end{tabular}
    \end{adjustbox}
    \caption{Average inference time(s) of a five cross-validation with the best hyperparameters. The \underline{\textbf{bold}} indicates the top result; - indicates that the model can not be applied to the task type; $\gg$ means the model can not produce results, due to issues such as GPU overflow or reaching a time limit, when dealing with some datasets.}
    \label{T.infer_time}
\end{table*}

\subsection{Interpretability}
In both practical applications and research, understanding the importance of individual features in machine learning is paramount, particularly when dealing with tabular data. This knowledge empowers decision-makers to enhance model performance and allocate resources judiciously. Significantly, in fields like finance and healthcare, where marginal improvements in predictions hold substantial real-world implications.

\begin{figure}[ht]
\centering
\subfigure[Feature importance of XGBoost.]{\includegraphics[width=0.45\textwidth]{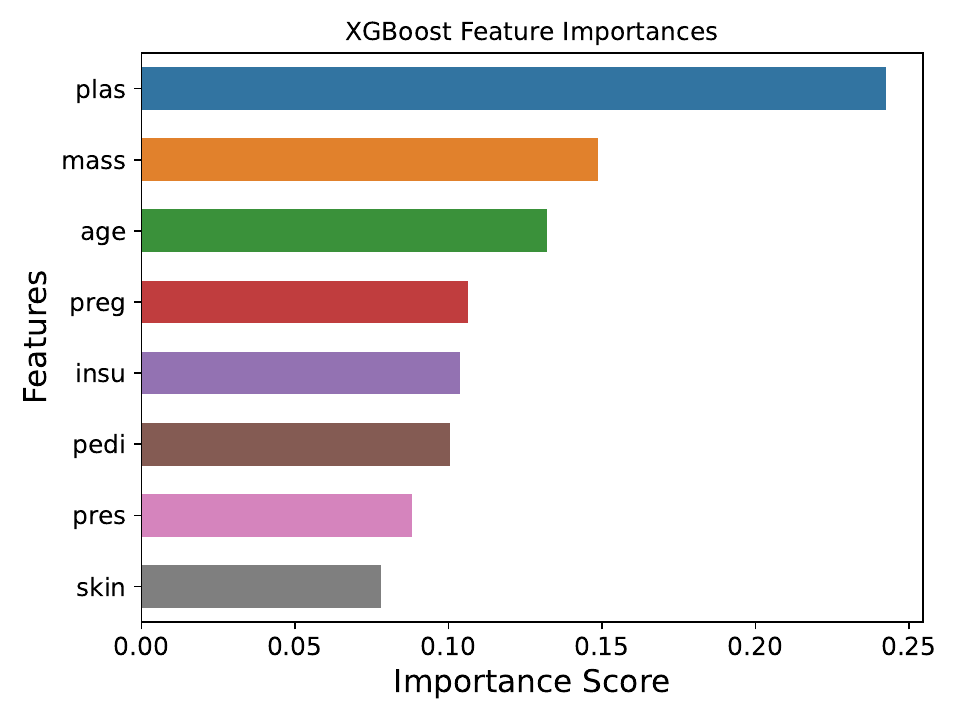}}
\subfigure[Feature importance of LightGBM.]{\includegraphics[width=0.45\textwidth]{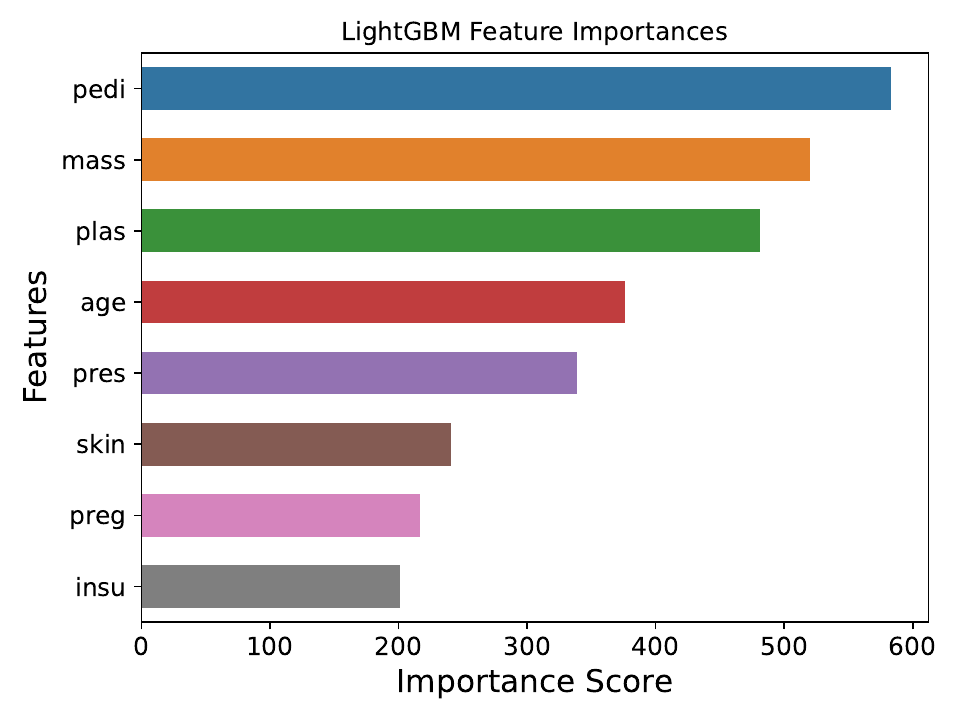}}
\subfigure[Feature importance of CatBoost.]{\includegraphics[width=0.45\textwidth]{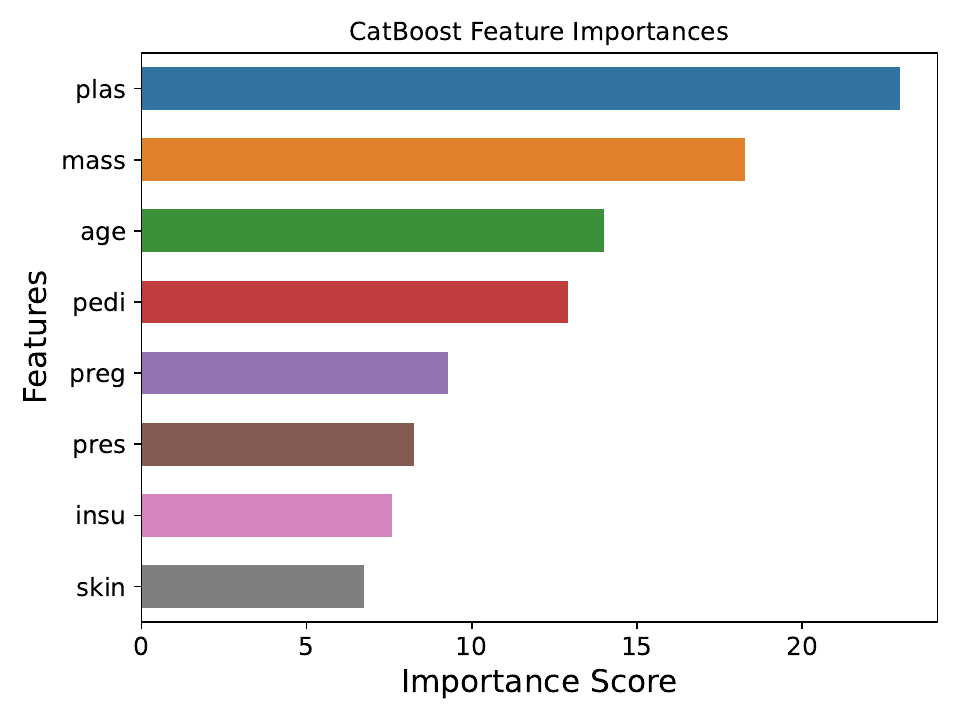}}
\subfigure[Feature importance of NCART.]{\includegraphics[width=0.45\textwidth]{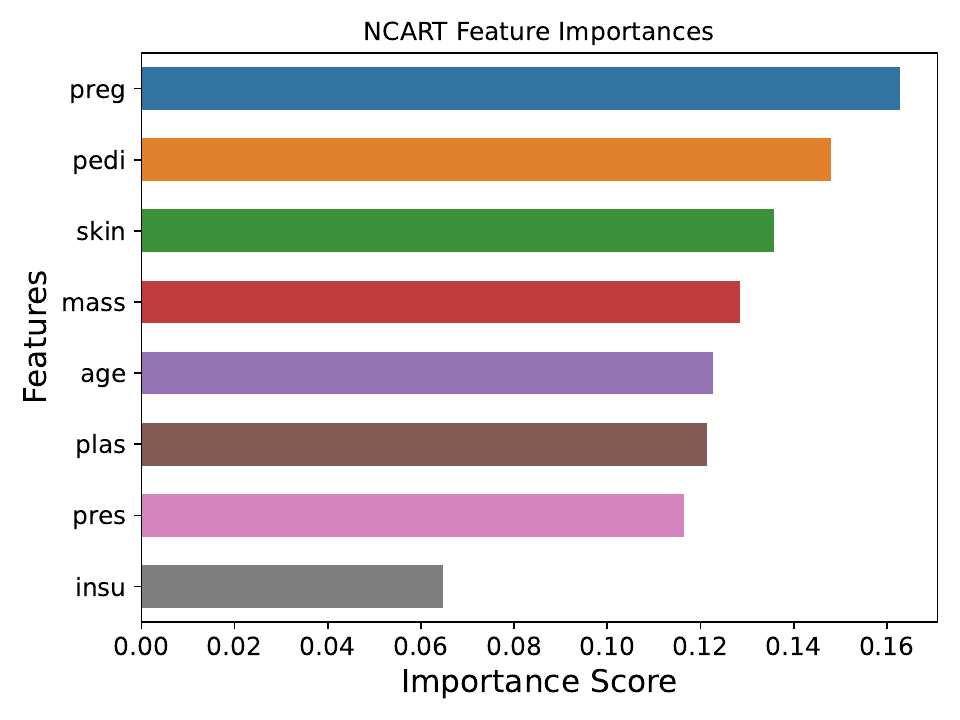}}
\caption{Feature importance of three boosting models and NCART on diabetes dataset.}
\label{f.feature_importance}
\end{figure}

In this subsection, we utilize four interpretable models, namely XGBoost, LightGBM, CatBoost, and NCART, to explore the feature importance in a medical dataset diabetes. Fig.~\ref{f.feature_importance} illustrates the importance scores obtained from these models, revealing variations in their assessment of feature importance.

These differences in feature importance primarily stem from the distinct structures and methodologies employed by each model. XGBoost, LightGBM, and CatBoost are ensemble methods based on GBDT, while the neural network operates on a different paradigm. The ensemble methods construct decision trees using diverse subsets of data and apply various randomization techniques, leading to discrepancies in their individual feature importance rankings. Additionally, the models' unique hyperparameters, such as learning rates, tree depths, and the number of trees, significantly influence the final importance scores for each feature. On the other hand, the neural network, with its layers, neurons, and activation functions, approaches feature importance from a different perspective.

Despite these differences in feature importance, all four models provide varying degrees of interpretability. This interpretability enables us to gain valuable insights into the factors driving their predictions, making them useful tools for understanding the relationships between features and target variables in the dataset.

\subsection{Limitations}

While NCART exhibits competitive performance in terms of numerical results compared to GBDT, there are several limitations to be considered:

\begin{itemize}
    \item NCART can achieve results that are competitive with GBDT, but there is still a noticeable performance gap, particularly when it comes to regression tasks. 
    \item NCART can compete with GBDT in terms of inference speed on small datasets, but its performance declines noticeably when working with larger datasets.
    \item One notable limitation of NCART is its inability to handle missing values. This can be a significant drawback in real-world datasets, where missing data is common, as it necessitates pre-processing steps to impute missing values before using NCART.
    \item NCART is more sensitive to hyperparameters in comparison to traditional tree models. Tuning hyperparameters becomes crucial to achieving optimal performance, but this process can be time-consuming and require expertise.
\end{itemize}

\section{Conclusions}
\label{s:conclu}

The performance of deep learning models in handling tabular data has been relatively restricted when compared to other machine learning techniques. Challenges such as limited data availability, high computational costs, and interpretability issues have posed significant obstacles to the widespread adoption of deep learning in real-life applications.

In this paper, we introduce NCART, a novel interpretable neural network that overcomes these challenges by incorporating differentiable decision trees within the ResNet architecture. By fusing decision trees with deep learning, our model combines the interpretability of decision trees with the powerful capabilities of deep neural networks.

NCART's simplicity in architecture makes it well-suited for datasets of varying sizes, while its efficiency leads to reduced computational expenses in comparison to state-of-the-art deep learning models. Our extensive numerical experiments reveal that NCART achieves competitive performance across datasets of different sizes, offering valuable insights into underlying patterns and relationships.

The introduction of NCART represents a significant step forward in addressing the limitations of deep learning in handling tabular data, opening up new possibilities for valuable applications in diverse scenarios. With its interpretability, efficiency, and performance, NCART paves the way for enhanced utilization of deep learning techniques in practical data analysis and decision-making tasks.



\section*{Acknowledgment}
This work was partially supported by the National Natural Science Foundation of China no. 12071190.

\appendix
\section{Sparse Function}
\label{a.equation}
We denote the $d$–probability simplex (the set of vectors representing probability distributions over $d$ choices) by $\Delta^d:= \{\mathbf{p} \in \mathbb{R}^d: \mathbf{p} \geq \mathbf{0}, \|\mathbf{p}\|_1 = 1\}$.

$sparsemax$ is an alternative to softmax which tends to yield sparse probability distributions:
\begin{equation*}
    sparsemax(\mathbf{z}) := argmin_{\mathbf{p}\in \Delta^d}\|\mathbf{p}-\mathbf{z}\|^2.
\end{equation*}

$entmax$ is a more general sparse function that has the following expression:
\begin{equation*}
    \alpha-entmax(\mathbf{z}) := argmin_{\mathbf{p}\in \Delta^d}\mathbf{p}^{\intercal}\mathbf{z} + H^T_{\alpha}(\mathbf{p}).
\end{equation*}
Here,  $H^T_{\alpha}(\mathbf{p})$ is called $Tsallis \alpha-entropie$ which is defined as:
\begin{align*}
    H^T_{\alpha}(\mathbf{p}) &= \left\{
    \begin{array}{ll}
        \frac{1}{\alpha(\alpha-1)}\sum_{j=1}^{d} (p_j-p_j^{\alpha}) & \alpha\neq 1  \\
        -\sum_{j=1}^{d} p_j \log p_j  & \alpha=1 ,
    \end{array}
    \right.
\end{align*}
where $p_j$ is the $j_{th}$ component of $\mathbf{p}$. It's worth noting that $1-entmax$ is equivalent to the $softmax$ function, and $2-entmax$ is equal to the $sparsemax$ function.

\section{Datasets Description}
\label{a.data}
Table. \ref{T.description} are the datasets used in this paper, the column \#Target means the number of distinct values in the label, and the column Objective gives a brief introduction to the objectives of each dataset. The selection of data follows the following principles: 1. Diversity of feature types; 2. Diversity of feature dimensions; 3. Diversity of sample numbers.

\begin{table*}[!ht]
\renewcommand\arraystretch{1.2}
\footnotesize
\setlength\tabcolsep{10pt}
    \centering
    \begin{adjustbox}{width=\textwidth}
    \begin{tabular}{lcccccc}
    \toprule[2pt]
        Dataset & \#Samples & \#Num.Feat. &\#Cat.Feat. & \#Target & Data id & Objective\\
    \midrule[1.5pt]
        \multicolumn{7}{c}{Binary Classification}\\
    \hline
        diabetes & 768  & 8 &0 & 2 & 37 &Patient's diabetes status recognition\\
    \hline
        credit-g & 1000  & 20  &13 & 2& 31 & People’s credit risks prediction\\
    \hline
        qsar-biodeg & 1055   & 41 &0 & 2 & 1494 & Analyze molecule biodegradation and structure\\
    \hline
        scene & 2407  & 299 &5 & 2& 312 & Scene recognition\\
    \hline
        ozone-level & 2534  & 72 &0 & 2&  1487 &Ozone level alarm forecasting\\
    \hline
        delta$\_$ailerons & 7129  & 5 &0 & 2&  803 & Aileron control prediction for aircraft\\
    \hline
        MagicTelescop & 13376  & 10 &0 & 2 &  44125 &Gamma signal recognition\\
    \hline
        jannis & 57580  & 54 &0 & 2 &  44131 & For challenging machine learning competitions\\
    \hline
        road-safety & 111762  & 32 &3 & 2 &  44161 & Personal injury road accidents recognition\\
    \hline
        Higgs & 940160  & 24 &0 & 2&  44129 &  Higgs boson signals recognition\\
    \midrule[1.5pt]
        \multicolumn{7}{c}{Multiclass Classification}\\
    \hline
        autoUniv-au7 & 700  & 12 &4 & 3 & 1553 & An advanced dataset generator for classification\\
    \hline
        plants-margin & 1600  & 64 &0 & 100 &  1491 & Plant leaf classification\\
    \hline
        waveform & 5000  & 40 &0 & 3 & 60 & Waves prediction\\
    \hline
        gas-drift & 13910  & 128 &0 & 6   & 1476 & Gas sensor drift prediction\\
    \hline
        EMNIST  & 131600  & 784 &0 & 47 &  41039 & Handwritten character digits recognition\\
    \midrule[1.5pt]
    \multicolumn{7}{c}{Regression}\\
    \hline
        analcatdata & 4052  & 7 &5 & 10  & 44055 & For analyzing categorical data\\
    \hline
        kin8nm & 8192  & 8 &0 & 8188  & 189 & Predict the dynamics of a robot arm\\
    \hline
        superconduct & 21263  & 79 &0 & 3007   & 44148 & The critical temperature prediction\\
    \hline
        house$\_$sales & 21613  & 17 &2 & 4028  & 44066 & House sale prices prediction\\
    \hline
        year & 515345  & 90 &0 & 89  & 44027 & Song release year prediction\\
    \bottomrule[2pt]
    \end{tabular}
    \end{adjustbox}
    \caption{Details of datasets (sorted by dataset size in each task type). \#Target in regression task means the number of unique values.}
    \label{T.description}
\end{table*}

\section{Optimization of hyperparameters}
\label{a:params}
Table.~\ref{T.hyperparam} lists the search range of hyperparameters, which refers to the survey \cite{borisov2022deep} and some adjustments are made on this basis to try to avoid GPU memory overflow and to lower the risk of overfitting on small-scale datasets. We implement the NCART model \footnote{The codes will be released to GitHub after acceptance.}  using PyTorch and employ the official open-source implementations for other models 
\footnote{XGBoost: \url{https://xgboost.readthedocs.io/en/stable/}}
\footnote{CatBoost: \url{https://catboost.ai/}}
\footnote{LightGBM: \url{https://lightgbm.readthedocs.io/en/latest/}}
\footnote{NODE: \url{https://github.com/Qwicen/node}}
\footnote{TabNet: \url{https://github.com/dreamquark-ai/tabnet}}
\footnote{SAINT: \url{https://github.com/somepago/saint}}
\footnote{FT-Transformer \& ResNet: \url{https://github.com/Yura52/rtdl}}
\footnote{TabCaps: \url{https://github.com/WhatAShot/TabCaps}}
\footnote{RLN: \url{https://github.com/irashavitt/regularization_learning_networks}}.

\begin{table*}[!ht]
\footnotesize
\centering
\begin{adjustbox}{width=\textwidth}
\begin{tabular}{ll|ll}
\toprule[2pt]
HyperParameters & Range & HyperParameters & Range \\
\midrule[1.5pt]
\multicolumn{4}{c}{XGBoost} \\
\hline
$num\_boost\_round$ & 1000  & $early\_stopping\_rounds$ & 10  \\
\hline
$max\_depth$ & LogUniformInt [2, 10] & $alpha$ & LogUniform [1e-8, 0.1]  \\
\hline
$lambda$ & LogUniform [0.5, 2] & $eta$ & LogUniform [0.05, 0.3]  \\
\midrule[1.5pt]

\multicolumn{4}{c}{CatBoost} \\
\hline
$iterations$ & 1000 & $od\_wait$ & 10\\
\hline
$max\_depth$ & LogUniformInt [2, 10] & $l2\_leaf\_reg$ & LogUniform [0.1, 2]\\
\hline
$learning\_rate$ & LogUniform [0.05, 0.3] &&\\
\midrule[1.5pt]

\multicolumn{4}{c}{LightGBM} \\
\hline
$iterations$ & 1000 & $early\_stopping\_round$ & 10\\
\hline
$num\_leaves$ & LogUniformInt [8, 64] & $lambda\_l_1$ & LogUniform [1e-8, 0.1]\\
\hline
$lambda\_l_2$ & LogUniform [1e-8, 0.1] & $learning\_rate$ & LogUniform [0.05, 0.3] \\
\midrule[1.5pt]

\multicolumn{4}{c}{NODE} \\
\hline
$num\_layers$& [2, 4, 8] & $total\_tree\_count$& [128, 256]  \\
\hline
$tree\_depth$& [4, 6, 8]  & $tree\_output\_dim$& [2, 3]  \\
\midrule[1.5pt]

\multicolumn{4}{c}{TabNet} \\
\hline
$n\_d$& LogUniform [8, 16] & $n\_steps$& UniformInt [1, 6]\\
\hline
$gamma$& Uniform [1, 2] & $n\_independent$& UniformInt [1, 5]\\
\hline
$n\_shared$& UniformInt [1, 5] & $momentum$& LogUniform [0.01, 0.4]\\
\hline
$mask\_type$& [sparsemax, entmax] && \\
\midrule[1.5pt]

\multicolumn{4}{c}{SAINT} \\
\hline
$dim$& [16, 32, 64] & $depth$& [1, 2, 3, 6]\\
\hline
$heads$& [2, 4, 8] & $dropout$& [0, 0.1, 0.2, 0.3, 0.4, 0.5]\\
\midrule[1.5pt]

\multicolumn{4}{c}{FT-Transformer} \\
\hline
$num\_blocks$& UniformInt [1, 6] & $num\_tokens$& [8, 16, 24, 32, 64, 128, 256, 512]\\
\hline
$dropout\_att$& [0, 0.1, 0.2, 0.3, 0.4, 0.5] & $dropout\_ffn$& [0, 0.1, 0.2, 0.3, 0.4, 0.5]\\
\hline
$dropout\_res$& [0, 0.1, 0.2, 0.3, 0.4, 0.5] & \\
\midrule[1.5pt]

\multicolumn{4}{c}{TabCaps} \\
\hline
$learning\_rate$& UniformInt [0.02, 0.1] & $num\_senior\_capsules$& UniformInt[1, 5]\\
\hline
$primary\_capsule\_size$& UniformInt[64, 128] & $num\_primary\_capsules$& UniformInt[4, 32]\\
\hline
$senior\_capsule\_size$& UniformInt[4, 32] & $num\_learnable\_words$& UniformInt[16, 64]\\
\midrule[1.5pt]

\multicolumn{4}{c}{RLN} \\
\hline
$layers$& UniformInt [2, 6] & $theta$& UniformInt [-12, -8]\\
\hline
$norm$& [1, 2] & & \\
\midrule[1.5pt]

\multicolumn{4}{c}{ResNet} \\
\hline
$n\_blocks$& UniformInt [1, 10] & $d\_hidden$& [32, 64, 128, 256, 512]  \\
\hline
$dropout$& [0, 0.1, 0.2, 0.3, 0.4, 0.5] && \\
\midrule[1.5pt]

\multicolumn{4}{c}{NCART} \\
\hline

$N$ in Eq. \eqref{e.ncart} & [8, 16, 32, 64] & $d$ in Eq. \eqref{e.feat_sel}& UniformInt [2, 10]  \\
\hline
$L$ in Eq. \eqref{e.feat_import3}& [2, 4]  & $h$ in Eq. \eqref{e.sel_mat}& \ [sparsemax, entmax]  \\
\bottomrule[2pt]
\end{tabular}
\end{adjustbox}
\caption{Hyperparameters space.}
\label{T.hyperparam}
\end{table*}

\section{Inference Time}
\label{a:infertime}
The average inference time(s) of a five cross-validation with the best hyperparameters are listed in Table \ref{T.infer_time_total}. 
\begin{table*}[!ht]
\renewcommand\arraystretch{1.4}
    \centering
    \begin{adjustbox}{width=\textwidth}
    \begin{tabular}{lccccccccccc}
    \toprule[2pt]
        Dataset & XGBoost & CatBoost &LightGBM & NODE & TabNet & SAINT& FTTrans & TabCaps& RLN & ResNet &  NCART \\
    \midrule[1.5pt]
    \multicolumn{12}{c}{Binary Classification}\\

    diabetes &  0.0039 &  0.0098 &  \underline{\textbf{0.0002}} &  0.0023 &  0.0070 &  0.4502 &  0.0033 &  0.0029 &  0.0854 &  0.0016 &  0.0014 \\
    
    credit-g &  0.0140 &  0.0077 &  \underline{\textbf{0.0002}} &  0.0042 &  0.0034 &  0.4367 & 0.0039 &  0.0029 &  0.0756 &  0.0016 &  0.0017 \\
    
    qsar-biodeg &  0.0042 &  0.0009 &  \underline{\textbf{0.0003}} &  0.0032 &  0.0071 &  0.4305 &  0.0121 &  0.0043 &  0.1015 &  0.0018 &  0.0015 \\
    
    scene   &  0.0045 &  0.0148 &  \underline{\textbf{0.0005}} &  0.1279 &  0.0156 &  $OOM$ & $OOM$  &  0.0071 &  0.0785 &  0.0042 &  0.0034 \\
    
    ozone-level &  0.0019 &  0.0018 &  \underline{\textbf{0.0003}} &  0.0131 &  0.0143 &  0.4957 &   0.0196 &  0.0032 &  0.0964 &  0.0037 &  0.0030 \\
    
    delta\_ailerons &  0.0020 &  0.0206 &  \underline{\textbf{0.0004}} &  0.0178 &  0.0266 &  0.4656 &    0.0266 &  0.0105 &  0.1564 &  0.0108 &  0.0086 \\
    
    MagicTelescop &  0.0033 &  0.0336 &  \underline{\textbf{0.0013}} &  0.0354 &  0.0196 &  0.5677 &       0.0398 &  0.0183 &  0.2344 &  0.0192 &  0.0181 \\
    
    jannis  &  0.0165 &  0.0211 &  \underline{\textbf{0.0057}} &  1.3936 &  3.0272 &  0.9445 &       0.2409 &  0.0716 &  0.5579 &  0.5238 &  0.0584 \\
    
    road-safety &  0.1075 &  0.1217 &  \underline{\textbf{0.0693}} &  2.1546 &  0.5648 &  1.1658 &       0.2741 &  0.1197 &  1.7843 &  0.1259 &  0.2867 \\
    
    Higgs  &  0.5332 &  \underline{\textbf{0.2298}} &  0.4163 &  4.0186 &  3.8084 &  6.5720 &       2.4771 &  1.4561 &  5.4842 &  2.3597 &  1.3039 \\

    \midrule[1pt]
    \multicolumn{12}{c}{Multiclass Classification}\\

        autoUniv-au7 &     0.0024 &  0.0154 & \underline{\textbf{ 0.0002}} &  0.0021 &  0.0031 &  0.3752 &       0.0037 &  0.0024 &  0.0983 &  0.0014 &  0.0029 \\
        
        plants-margin &  0.0130 &  0.0276 &  0.0221 &  0.1441 &  0.0119 &  0.2342 &       0.0055 &  0.0105 &  0.0793 &  0.0034 &  \underline{\textbf{0.0022}} \\
        
        waveform  &  0.0061 &  0.0041 &  \underline{\textbf{0.0012}} &  0.0126 &  0.0061 &  0.4448 &       0.0098 &  0.0067 &  0.1406 &  0.0244 &  0.0057 \\
        
        gas-drift  &  0.0116 &  0.0166 &  \underline{\textbf{0.0046}} &  0.0492 &  0.0281 &  0.9341 &       0.1340 &  0.0160 &  0.2388 &  0.0219 &  0.0209 \\
        
        EMNIST  &  2.2039 &  1.0328 &  5.3717 &  2.3124 &  3.0943 &  $OOM$ &  $OOM$ &  0.3663 &  1.2044 &  \underline{\textbf{0.2330}} &  0.7222\\

    \midrule[1pt]
    \multicolumn{12}{c}{Regrssion}\\
        
        analcatdata  & 0.0033 &  0.0062 &  \underline{\textbf{0.0002}} &       0.0294 &   0.0269 &   0.4621 &  0.0256 & -& 0.1183 &  0.0071 &  0.0067 \\

        kin8nm & 0.0083 &  0.0052 &  \underline{\textbf{0.0045}} &       0.0121 &   0.0099 &   0.4858 &  0.0282 & -& 0.0649 &  0.0110 &  0.0096 \\

        superconduct & \underline{\textbf{0.0139}} &  0.0445 &  0.0249 &       0.2725 &   0.3562 &   0.7072 &  0.1020 & -& 0.0938 &  0.0316 &  0.0316\\
        
        house$\_$sales & 0.0072 &  0.0159 &  \underline{\textbf{0.0032}} &       0.5453 &   0.0812 &   0.6169 &  0.1300 & -& 0.1115 &  0.0255 &  0.0759\\

        year & 0.7122 &  \underline{\textbf{0.2633}} &  0.5768 &  $TimeOut$ &  11.2118 &  10.7570 &  1.2313 & -& 5.7785 &  2.1037 &  0.7257 \\

    \bottomrule[2pt]
    \end{tabular}
    \end{adjustbox}
    \caption{Average inference time(s) of a five cross-validation with the best hyperparameters.  The \underline{\textbf{bold}} indicates the top result; - indicates that the model can not be applied to the task type; $OOM$ represents there exists GPU overflow; $TimeOut$ means the running time exceeds the time limit.}
    \label{T.infer_time_total}
\end{table*}

\section{Training Time}
\label{a:traintime}
The detailed training time(s) of five cross-validation with the best hyperparameters are shown in Table \ref{T.running_time}. 
\begin{table*}[!ht]
\renewcommand\arraystretch{1.4}
    \centering
    \begin{adjustbox}{width=\textwidth}
    \begin{tabular}{lccccccccccc}
    \toprule[2pt]
        Dataset & XGBoost & CatBoost &LightGBM & NODE & TabNet & SAINT& FTTrans& TabCaps & RLN & ResNet &  NCART \\
    \midrule[1.5pt]
    \multicolumn{12}{c}{Binary Classification}\\

        diabetes &    \underline{\textbf{0.08}} &      1.08 &     0.11  &     32.85 &     0.13 &    156.10 &         1.09 & 0.74 &  24.22 &     1.99 &     5.79\\

        credit-g  &     0.18 &      4.19 &      0.16 &     84.10 &   \underline{\textbf{0.07}} &     73.99 &          6.28 & 0.39 &  16.08 &     2.44 &     7.18 \\

        qsar-biodeg & \underline{\textbf{0.13}} &      3.21 &      0.18 &     20.84 &     0.14 &    199.24 &        29.71 & 2.02&  46.89 &     1.39 &    13.34 \\

        scene & 0.44 &     3.48 &      \underline{\textbf{0.21}} &    193.95 &    14.73 &  $OOM$ &       $OOM$ & 2.13&  34.73 &     1.63 &    13.16\\

        ozone-level &   \underline{\textbf{0.11}} &      1.91 &      0.24 &    211.53 &     2.15 &    298.27 &        70.45 & 1.30&    4.98 &     3.23 &     8.26 \\

        delta$\_$ailerons  &    \underline{\textbf{0.06}} &      2.95 &      0.17 &    346.49 &    50.27 &     89.17 &         91.79 & 1.81&   5.11 &    10.97 &    15.22\\

        MagicTelescop  &    \underline{\textbf{0.29}} &     10.61 &      0.38 &    369.31 &    31.19 &    180.62 &          18.59 & 22.92&  59.22 &    20.43 &    19.19\\

        jannis &     1.31 &     18.75 &     \underline{\textbf{0.86}} &    500.89 &   684.32 &    169.47 &         155.88 & 37.78&  83.88 &    48.70 &    36.45 \\
        
        road-safety &    3.87 &      5.85 &     \underline{\textbf{3.82}} &   1000.29 &  1306.03 &    443.30 &         805.73 & 135.45&  25.47 &   192.86 &   132.75  \\

        Higgs  &    \underline{\textbf{6.63}} &     9.63 &    10.73 &    954.56 &  8209.51 &   1162.72 &        2914.61 & 1022.71&  599.21 &  1015.46 &  1424.64\\

    \midrule[1pt]
    \multicolumn{12}{c}{Multiclass Classification}\\

        autoUniv-au7 &     0.20 &      1.01 &      0.18 &     30.11 &     \underline{\textbf{0.06}} &    136.29 &          7.06 & 0.33&  41.37 &     1.11 &     4.13\\

        plants-margin &    6.53 &     10.05 &   6.66 &    405.50 &     81.59 &    103.81 &           39.45 & 27.11&   4.60 &     \underline{\textbf{3.94}} &     4.85 \\

        waveform &     1.19 &      2.30 &      \underline{\textbf{0.45}} &     28.52 &     9.57 &    129.35 &          4.24 & 2.38 & 46.18 &     4.34 &     9.48\\

        gas-drift &     \underline{\textbf{2.10}} &      6.68 &     6.81 &   1145.76 &    41.46 &    455.32 &         131.10 & 8.73&  33.51 &    14.81 &    35.15  \\
        
        EMNIST &   251.56 &    207.63 &    249.05 &   2649.74 &  1536.03 &  $OOM$ &  $OOM$ & 304.60&  \underline{\textbf{98.74}} &    99.62 &   288.04 \\
    \midrule[1pt]
    Average Rank & \underline{\textbf{1.93}} & 3.93 & 2.20 & 9.53 & 6.73 & 10.13 & 8.20 &4.80 & 7.13 & 5.07 & 6.47 \\

    Best/Worst& \underline{\textbf{7/0}} & 0/0& 4/0 & 0/4 & 2/3 & 0/9 & 0/3 & 0/0 & 1/0 & 1/0 & 0/0\\

    Total time & \underline{\textbf{274.67}}   &   289.33 & 280.00   &  7974.43    &  11967.27  &  $\gg$4420.14  &  $\gg$4276.10   &   1570.42   &   1164.22   &   1422.93 &1991.12 \\
    \bottomrule[2pt]
    \bottomrule[2pt]
    \multicolumn{12}{c}{Regression}\\
        
        analcatdata  &     0.31 &      1.10 &      \textbf{0.15} &     97.77 &     7.66 &     42.17 &           9.39 & -&  30.32 &     6.41 &    11.49 \\

        kin8nm &     1.69 &      4.13 &     \textbf{1.43} &    210.32 &   108.63 &    323.10 &          29.08 & -&   39.24 &    15.71 &    14.98 \\

        superconduct &     \textbf{2.46} &     13.37 &     16.24 &    858.54 &   132.95 &    569.11 &         279.68 & -&   81.50 &    28.59 &    44.02\\
        
        house$\_$sales &    0.81 &     5.07 &    \textbf{0.64} &   1744.48 &   132.61 &    387.82 &          70.98 & -&  169.44 &    22.45 &    36.68 \\

        year & 14.22 &     \textbf{8.25} &   12.92 &  $TimeOut$ &  3036.32 &   3916.41 &        1805.23 & -&  120.02 &  1594.27 &   690.45\\

    \midrule[1pt]
    Average Rank & 2.0 & 2.4 & \underline{\textbf{1.6}} & 9.6 & 8.0 & 8.6 & 6.8 & -&  6.2 & 4.8 & 5.0 \\

    Best/Worst& 1/0 & 1/0& \underline{\textbf{3/0}} & 0/3 & 0/1 & 0/1 & 0/0 & -& 0/0 & 0/0 & 0/0\\

    Total time & \underline{\textbf{19.48}}   &    31.91 & 31.37   &  $\gg$2911.10    &  3580.22  &  5238.61  &  2686.11   & -&   401.14   &   1667.43   &   789.83 \\
    \bottomrule[2pt]
    \end{tabular}
    \end{adjustbox}
    \caption{Average training time(s) of a five cross-validation with the best hyperparameters. The \underline{\textbf{bold}} indicates the top result; - indicates that the model can not be applied to the task type; $OOM$ represents there exists GPU overflow; $TimeOut$ means the running time exceeds the time limit.}
    \label{T.running_time}
\end{table*}

\bibliographystyle{elsarticle-num} 
\bibliography{references}

\end{document}